%% file: neurips_2026.tex
\documentclass{article}

\PassOptionsToPackage{authoryear, round}{natbib}
\usepackage[preprint]{neurips_2026}

\usepackage[utf8]{inputenc} % allow utf-8 input
\usepackage[T1]{fontenc}    % use 8-bit T1 fonts
\usepackage{hyperref}       % hyperlinks
\usepackage{url}            % simple URL typesetting
\usepackage{booktabs}       % professional-quality tables
\usepackage{amsfonts}       % blackboard math symbols
\usepackage{nicefrac}       % compact symbols for 1/2, etc.
\usepackage{microtype}      % microtypography
\usepackage{xcolor}         % colors

\usepackage{amsmath}
\usepackage{amsthm}
\usepackage{graphicx} 
\usepackage{float}
\usepackage{subcaption}
\usepackage{wrapfig}
\usepackage{multirow}
\usepackage{arydshln}
\usepackage{enumitem}
\usepackage{placeins}

\usepackage[most]{tcolorbox}

\tcbset{
  mytheoremstyle/.style={
    enhanced,
    breakable,
    width=\linewidth,
    center,
    colback=gray!7,
    colframe=gray!7,
    coltitle=black,
    coltext=black,
    fonttitle=\bfseries,
    fontupper=\normalfont,
    boxrule=0pt,
    arc=2pt,
    outer arc=2pt,
    left=8pt,
    right=8pt,
    top=2pt,
    bottom=6pt,
    before skip=8pt,
    after skip=8pt,
    fontupper=\itshape,
  }
}

\newtcbtheorem[]{definition}{Definition}
{mytheoremstyle}{def}

\newtcbtheorem[]{proposition}{Proposition}
{mytheoremstyle}{propos}

\newtcbtheorem[]{property}{Property}
{mytheoremstyle}{pty}
\title{Deterministic Decomposition of Stochastic Generative Dynamics}

\author{
Xingyu Song$^{1}$ \quad
Yuan Mei$^{1,2}$\thanks{Visiting student at The University of Tokyo from Zhejiang University.} \quad
Naoya Takeishi$^{1}$\\
$^{1}$The University of Tokyo\\
$^{2}$Zhejiang University\\
\texttt{\{xsong,umegen,ntake\}@g.ecc.u-tokyo.ac.jp}
}

\begin{document}

\maketitle

\begin{abstract}

Modern generative models can be understood as probability transport from a simple base distribution to a target data distribution.
Deterministic transport models offer tractable velocity-field parameterizations, whereas stochastic generative models capture richer density evolution through drift and diffusion.
Yet when stochastic dynamics are described through deterministic velocity fields, the effects of drift and diffusion are often compressed into a single effective field, obscuring the distinct roles of deterministic evolution and stochastic fluctuation.
In this work, we show that the deterministic field \(b_t\) of a stochastic generative process admits a natural transport--osmotic decomposition that separates deterministic transport from stochastic, diffusion-induced effects:
\(b_t = u_t + d_t\), where \(u_t\) governs marginal probability transport and \(d_t\) captures an osmotic effect induced by diffusion and determined by the marginal score.
Based on this decomposition, we propose Bridge Matching, a flow-based framework for learning decomposed generative dynamics through both marginal and conditional formulations.
In generative modeling experiments, we recombine the learned components as \(b_t = u_t + \lambda_d d_t\), showing that the proposed decomposition enables interpretable and controllable sampling by adjusting the osmotic contribution in probability transport.
The implementation is publicly available \href{https://github.com/xingyu-song/bridge_matching}{here}.

\end{abstract}

%%%%%%%%%%%%%%%%%%%%%%%%%%%%%%%%%%%%%%%%%%%%%%%%%%%%%%%%%%%%%%%%%%
%%%%%%%%%%%%%%%%%%%%%%%%%%%%%%%%%%%%%%%%%%%%%%%%%%%%%%%%%%%%%%%%%%
\section{Introduction}
\label{introduction}
\vspace{-3pt}
Generative modeling has become a central paradigm in modern machine learning, enabling the synthesis of high-dimensional data across domains such as vision \citep{song2024animationbasedaugmentationapproachaction, esser2024scaling}, audio \citep{liu2023audioldm, evans2024stable}, and scientific simulation \citep{gao2024generative, song2024quatergcnenhancing3dhuman}.
A unifying view of modern generative modeling is \textbf{probability transport} \citep{leonard2014survey, lipman2023flow, albergo2023interpolants, liu2023rectified}: samples are generated by evolving a simple base distribution \(\pi_0\) into a target data distribution \(\pi_1\). 

A key distinction among these approaches lies in the dynamics governing the evolution of marginal distributions.
In deterministic formulations, samples evolve according to an ordinary differential equation (ODE) defined by a velocity field \citep{chen2018neural, grathwohl2019ffjord}, so that the associated marginal density evolves according to the continuity equation \citep{villani2009optimal}.
This makes the learned dynamics relatively tractable: the velocity field directly characterizes the transport of probability mass over time, and can be learned by regression-based objectives \citep{lipman2023flow,tong2023minibatchot}.
In stochastic formulations, by contrast, samples evolve according to stochastic differential Equations (SDEs), while their marginal densities are governed by the Fokker--Planck equation
\citep{anderson1982reverse,risken1996fokker,song2021score}.
Here, density evolution is determined jointly by drift and diffusion, making stochastic dynamics expressive but also more difficult to interpret and control.

Nevertheless, deterministic and stochastic models should not be viewed as separate families.
Instead, several existing perspectives connect stochastic generative dynamics to deterministic dynamics or deterministic representations.
For example, probability flow ODEs define deterministic dynamics whose marginal distributions match those of the corresponding stochastic diffusion process
\citep{song2021score,chen2023probability}.
Stochastic interpolants and Schr\"odinger-bridge-based methods further emphasize pathwise and bidirectional structures, connecting endpoint distributions through stochastic or deterministic interpolating dynamics
\citep{albergo2023interpolants,chen2022sb, leonard2014survey, de2021diffusion}.
Related deterministic transport formulations construct velocity fields that realize prescribed probability paths through the continuity equation
\citep{lipman2023flow,liu2023rectified}.
Other works improve or regularize learned generative trajectories through path straightening or curvature-based criteria
\citep{improveRF2024,lee2023curvature}.
Together, these connections show that stochastic and deterministic generative models are closely related at the level of marginal evolution.

However, existing deterministic reformulations of stochastic dynamics typically represent marginal evolution with a single effective velocity field. 
Although this enables deterministic sampling and efficient training, it collapses the mechanisms underlying stochastic dynamics into one representation. 
In particular, probability-mass transport and diffusion-induced density reshaping are absorbed into the same marginal velocity, making it difficult to analyze, control, or recombine these effects independently. 
Consequently, even when deterministic dynamics reproduce the same marginal distributions, the contribution of stochasticity remains hidden.

\begin{figure}[t]
\vspace{0pt}
\centering
\begin{minipage}[b]{0.42\linewidth}
  \vspace{0pt}
  \centering
  \includegraphics[width=\linewidth]{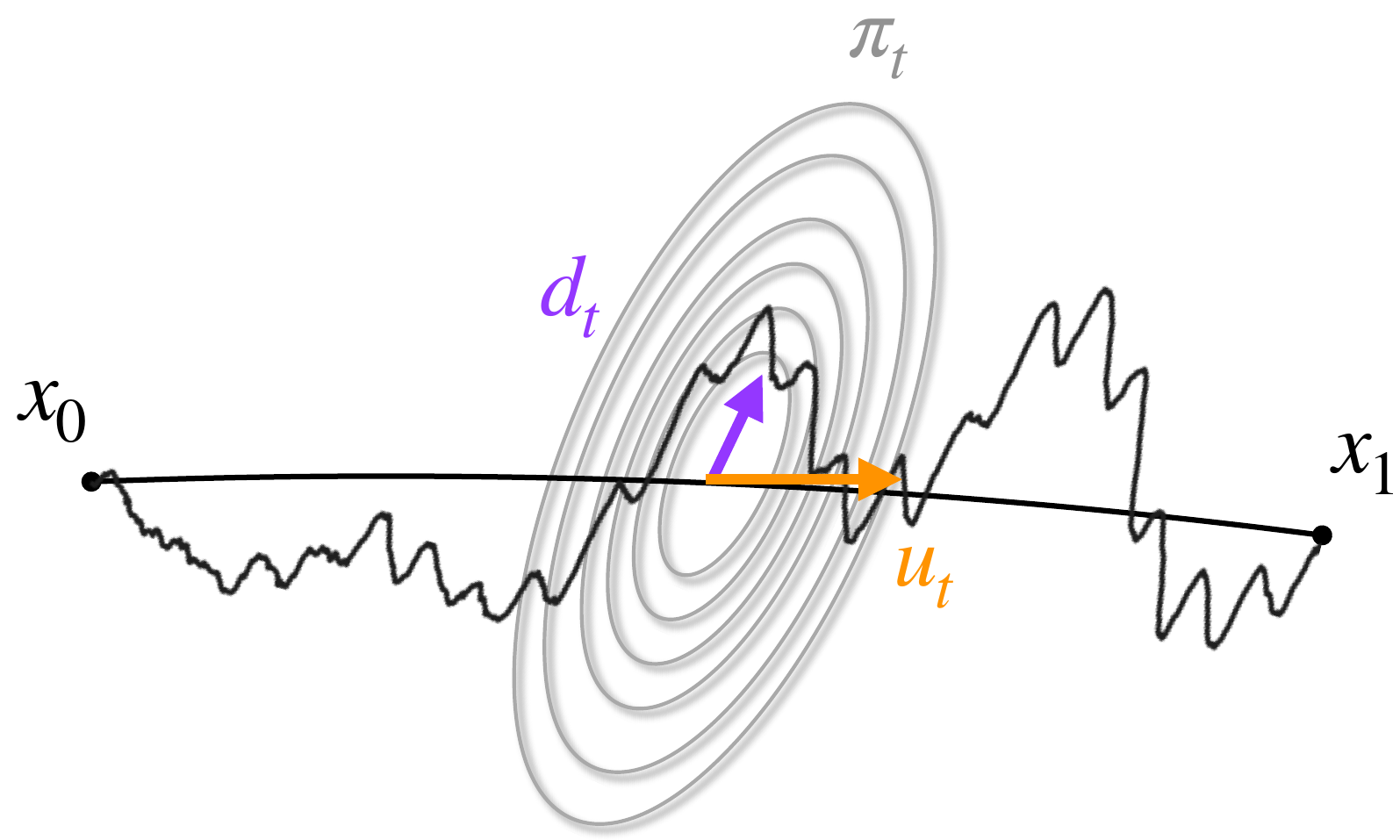}
\end{minipage}
\hfill
\begin{minipage}[b]{0.55\linewidth}
  \vspace{0pt}
  \centering
  \caption{Overview of the proposed Transport--Osmotic decomposition. Dynamics from source sample \(x_0\) to target sample \(x_1\) are decomposed into transport field \(u_t\) and osmotic field \(d_t\) over the marginal distribution \(\pi_t\). Smooth and noisy curves denote deterministic and stochastic trajectories.}
  \label{fig:overview}
\end{minipage}
\vspace*{-1em}
\end{figure}

In this work, we show that stochastic generative dynamics admit a more structured deterministic representation.
Our key observation is that stochastic processes exhibit an intrinsic bidirectional structure: the forward and backward drift fields associated with the same marginal path are coupled through the marginal score.
This coupling naturally induces a \textbf{Transport--Osmotic decomposition} into two complementary deterministic vector fields as shown in Fig.~\ref{fig:overview}.
The transport field governs the movement of probability mass and determines marginal evolution through a continuity equation, whereas the osmotic field captures diffusion-induced, density-dependent effects.
In this view, stochasticity is not treated merely as an external noise source, but is represented as a structured component of deterministic marginal dynamics.

Building on this decomposition, we propose \textit{Bridge Matching}, a framework for learning decomposed generative dynamics
Rather than learning a single effective velocity field directly, Bridge Matching learns separate transport and osmotic (density-shaping) fields that can be recombined to construct forward or backward generative dynamics.
We develop both marginal and conditional formulations: 
the marginal formulation follows the distribution-level definition of the decomposition, while the conditional formulation provides a tractable implementation based on conditional probability paths with analytically available scores.
This enables the decomposition to be learned with regression-style objectives similar to Flow Matching, while preserving the structural distinction between transport and diffusion-induced effects.

% \begin{wrapfigure}[16]{r}{0.44\linewidth}
% \vspace{0pt}
% \centering
% \includegraphics[width=0.90\linewidth]{fig/res/overview.pdf}
% \caption{
% Overview of the proposed Transport--Osmotic decomposition.
% Dynamics from source sample \(x_0\) to target sample \(x_1\) are decomposed into transport field \(u_t\) and osmotic field \(d_t\) over the marginal distribution \(\pi_t\).
% Smooth and noisy curves denote deterministic and stochastic trajectories.
% }
% \label{fig:overview}
% \vspace{-10pt}
% \end{wrapfigure}

% NOTE: wrapfig + itemize side by side looks a bit strange I think

\paragraph{Contributions.}
\begin{enumerate}[leftmargin=1.2em, label=\arabic*., itemsep=0em, topsep=0em]
    \item We derive a Transport--Osmotic decomposition of stochastic generative dynamics, showing that deterministic and stochastic models share a common marginal transport backbone while differing in how diffusion-induced effects enter the sample dynamics.
    \item We propose \emph{Bridge Matching}, a training framework that learns the transport and osmotic fields through marginal and conditional formulations, connecting score-based information with flow-style regression objectives.
    \item We empirically show that recombining the learned fields improves generation performance and enables controllable, interpretable sampling by adjusting the relative strength of transport and density-dependent effects.
\end{enumerate}

%%%%%%%%%%%%%%%%%%%%%%%%%%%%%%%%%%%%%%%%%%%%%%%%%%%%%%%%%%%%%%%%%%
%%%%%%%%%%%%%%%%%%%%%%%%%%%%%%%%%%%%%%%%%%%%%%%%%%%%%%%%%%%%%%%%%%
\section{Generative modeling via probability transport}
\label{generative_modeling_via_probability_transport}

In the data space \(\mathcal{X} \subseteq \mathbb{R}^d\), a generative model defines a parametric probability distribution
 \(p_\theta(x)\) with learnable parameters \(\theta \in \Theta\).
The goal is to approximate an unknown data distribution \(p_{\text{data}}(x)\) such that \(p_\theta(x) \approx p_{\text{data}}(x)\).

A unifying perspective formulates generative modeling as a probability transport problem \citep{villani2009optimal,leonard2014survey}, 
where samples from a simple base distribution \(\pi_0\) are progressively transformed into samples from a target data distribution \(\pi_1\)
 \citep{rezende2015nf,sohl2015diffusion,lipman2023flow}.
This transformation is described by a continuous family of intermediate distributions \(\{\pi_t\}_{t \in [0,1]}\), with boundary conditions \(\pi_0\) and \(\pi_1\) at \(t=0\) and \(t=1\).
From this perspective, generative modeling paradigms differ in how they specify the probability path and in the sample-level dynamics used to realize it. 
More specifically, these dynamics may be deterministic, as in continuous normalizing flows and flow matching, or stochastic, as in diffusion and score-based models
 \citep{chen2018neural,song2021score, grathwohl2019ffjord}.

%%%%%%%%%%%%%%%%%%%%%%%%%%%%%%%%%%%%%%%%%%%%%%%%%%%%%%%%%%%%%%%%%%
\subsection{Deterministic transport and the continuity equation}
\label{Deterministic_Transport_and_the_Continuity_Equation}
Following the probability transport view, deterministic transport models realize a distribution-level path \(\{\pi_t\}_{t\in[0,1]}\) through sample-level deterministic dynamics.
Here, \(\pi_t\) denotes the marginal density of \(X_t\) at time \(t\), satisfying \(\int \pi_t(x)\,dx = 1\) for all \(t \in [0,1]\).
In deterministic transport, each initial sample \(X_0 \sim \pi_0\) is mapped to \(X_t = \phi_t(X_0)\) by a flow map \(\phi_t\), which pushes forward the initial distribution as \(\pi_t = [\phi_t]_\# \pi_0\).
The flow map is generated by a time-dependent velocity field \(v : [0,1]\times\mathbb{R}^d \to \mathbb{R}^d\).
This velocity field governs the sample-level dynamics through the ODE \citep{chen2018neural, grathwohl2019ffjord}: 
\begin{equation}
\label{eq:ode_for_fm}
    \frac{dX_t}{dt}= v_t(X_t).
\end{equation}
Although the ODE describes the motion of individual samples, it also induces an evolution of the marginal density \(\pi_t\). Since probability mass is transported by the deterministic velocity field and is neither created nor destroyed, \(\pi_t\) satisfies the continuity equation: 
\begin{equation}
\label{eq:continuity_equation}
    \partial_t\pi_t(x) + \nabla\cdot(\pi_t(x)v_t(x)) = 0. 
\end{equation}
This equation states that local changes in density are determined by the divergence of the probability flux \(\pi_t v_t\).
In other words, probability mass changes at a location only because mass flows into or out of that location under the velocity field.

%%%%%%%%%%%%%%%%%%%%%%%%%%%%%%%%%%%%%%%%%%%%%%%%%%%%%%%%%%%%%%%%%%
\subsection{Stochastic transport and the Fokker--Planck Equation}
\label{stochastic_transport_and_the_fokker_planck_equation}
In contrast to deterministic transport, stochastic transport realizes a probability path through sample-level stochastic dynamics, where each sample evolves under both a drift field and random diffusion noise.
Such dynamics are commonly described by an SDE of the form: 
\begin{equation}
\label{eq:sde_for_diffusion}
    dX_t = b_t(X_t)dt + \sqrt{\beta_t} dW_t.
\end{equation}
Here, \(b:[0,1] \times\mathbb{R}^d \to \mathbb{R}^d\) is the drift field that transports mass directionally. 
\(\beta: [0,1] \to\mathbb{R}_{>0}\) is the time-dependent diffusion coefficient controlling the noise scale.
\(dW_t\) is the stochastic differential which denotes the infinitesimal increment of Brownian motion \citep{oksendal2003stochastic}, injecting random perturbations.
Compared with an ODE, the additional Brownian term makes each trajectory random even when the initial condition is fixed.

While deterministic ODE dynamics induce the continuity equation, stochastic SDE dynamics induce the Fokker-–Planck equation:
\begin{equation}
\label{eq:fokker_plank}
    \partial_t \pi_t(x)
    =
    -\nabla\cdot(\pi_t(x) b_t(x))
    +
    \frac{\beta_t}{2}\Delta \pi_t(x).
\end{equation}
The first term is a transport term, analogous to the continuity equation, while the second term is a diffusion term that spreads probability mass according to local density geometry.
Therefore, stochastic transport changes the marginal density not only by moving mass through the drift field, but also by reshaping it through diffusion.

%%%%%%%%%%%%%%%%%%%%%%%%%%%%%%%%%%%%%%%%%%%%%%%%%%%%%%%%%%%%%%%%%%
%%%%%%%%%%%%%%%%%%%%%%%%%%%%%%%%%%%%%%%%%%%%%%%%%%%%%%%%%%%%%%%%%%
\section{Deterministic decomposition of stochastic generative dynamics}
\label{Deterministic_Decomposition_of_Stochastic_Generative_Dynamics}
In this section, we derive a Transport--Osmotic decomposition from the intrinsic bidirectional structure of stochastic dynamics, separating marginal probability transport from diffusion-induced density-dependent effects.
Throughout this section, we assume that the marginal densities \(\pi_t\) are smooth and strictly positive on their support, with sufficient regularity to justify the integrations by parts and differentiations below.
%%%%%%%%%%%%%%%%%%%%%%%%%%%%%%%%%%%%%%%%%%%%%%%%%%%%%%%%%%%%%%%%%%
\subsection{Bidirectional structure of stochastic dynamics}

Stochastic processes described by SDEs admit an intrinsic bidirectional structure.
Under suitable regularity conditions, the same marginal density path can be represented by both forward-time and reverse-time dynamics \citep{anderson1982reverse,haussmann1986reverse}.
A canonical example is a diffusion process, whose forward and backward representations share the same family of marginal densities but generally have different drift fields.

For simplicity, consider a diffusion process with constant scalar diffusion coefficient \(\beta>0\) and marginal density \(\pi_t\).
Its forward-time dynamics are given by
\begin{equation}
\label{eq:forward_sde}
dX_t = b_t^\rightarrow(X_t)\,dt + \sqrt{\beta}\,dW_t,
\end{equation}
where \(b_t^\rightarrow\) denotes the forward drift and \(W_t\) is a standard Brownian motion.
The same stochastic process, viewed in reverse time, admits a reverse-time representation.
Let \(\tau=1-t\) and define \(Y_\tau=X_{1-\tau}\). 
The reverse-time dynamics can be written as
\begin{equation}
\label{eq:reverse_sde}
dY_\tau = \tilde b_\tau(Y_\tau)\,d\tau + \sqrt{\beta}\,d\bar{W}_\tau,
\end{equation}
where \(\tilde b_\tau\) is the reverse-time drift and \(\bar{W}_\tau\) is Brownian motion under the reversed filtration.
Throughout this work, we express the backward drift in the original forward-time orientation by defining
\(b_t^\leftarrow(x) := -\tilde b_{1-t}(x).\)

Although the forward and reverse processes describe opposite time directions, they are consistent with the same marginal density path \(\{\pi_t\}_{t\in[0,1]}\).
The forward dynamics induce the usual Fokker--Planck equation:
\begin{equation}
\label{eq:forward_fpe}
\partial_t \pi_t(x)
=
-\nabla\cdot\bigl(\pi_t(x)b_t^\rightarrow(x)\bigr)
+
\frac{\beta}{2}\Delta \pi_t(x).
\end{equation}
Using the backward drift expressed in the original forward-time orientation, the corresponding marginal equation takes the form:
\begin{equation}
\label{eq:backward_fpe}
\partial_t \pi_t(x)
=
-\nabla\cdot\bigl(\pi_t(x)b_t^\leftarrow(x)\bigr)
-
\frac{\beta}{2}\Delta \pi_t(x).
\end{equation}
Comparing these two equations and using
\(\Delta \pi_t = \nabla\cdot(\pi_t \nabla \log \pi_t)\) yields
\begin{equation}
\label{eq:bidirectional_consistency_general}
b_t^\rightarrow(x) - b_t^\leftarrow(x)
=
\beta \nabla \log \pi_t(x) + h_t(x),
\qquad
\nabla\cdot(\pi_t h_t)=0.
\end{equation}
Here, \(h_t\) is a density-weighted divergence-free component that does not affect the marginal evolution.
Throughout this work, we focus on the canonical gradient representative, for which \(h_t=0\), giving
\begin{equation}
\label{eq:bidirectional_consistency}
b_t^\rightarrow(x) - b_t^\leftarrow(x)
=
\beta \nabla \log \pi_t(x).
\end{equation}

This relation shows that the forward and backward drifts are not independent.
Their discrepancy is determined by the marginal score \(\nabla \log \pi_t\), which encodes the diffusion-induced effect of stochastic dynamics at the level of marginal evolution.
Thus, even though individual trajectories are stochastic, the effect of stochasticity on the marginal path can be represented through a deterministic, density-dependent vector field.

%%%%%%%%%%%%%%%%%%%%%%%%%%%%%%%%%%%%%%%%%%%%%%%%%%%%%%%%%%%%%%%%%%
\subsection{Transport--Osmotic decomposition}

The bidirectional structure of stochastic dynamics, characterized by the score-based coupling between forward and backward drifts, naturally induces a symmetric--antisymmetric decomposition.
In the following, we use \(b_t^\rightarrow\) and \(b_t^\leftarrow\) to denote the forward and backward SDE drifts, and adopt the canonical representative by setting density-weighted divergence-free components to zero. 

\begin{definition}{}{decomp}
Given forward and backward drift fields \(b_t^\rightarrow\) and \(b_t^\leftarrow\) associated with the same marginal path \(\{\pi_t\}_{t\in[0,1]}\), we define the symmetric component \(u_t\) and antisymmetric component \(d_t\) as
\begin{equation}
\label{eq:u}
  u_t(x) := \frac{1}{2}\bigl(b_t^\rightarrow(x) + b_t^\leftarrow(x)\bigr),
\end{equation}
\begin{equation}
\label{eq:d}
  d_t(x) := \frac{1}{2}\bigl(b_t^\rightarrow(x) - b_t^\leftarrow(x)\bigr).
\end{equation}
\end{definition}

Correspondingly, the original forward and backward drifts can be reconstructed as
\begin{equation}
\label{eq:forward_backward_construction}
  b_t^\rightarrow(x) = u_t(x) + d_t(x),
  \qquad
  b_t^\leftarrow(x) = u_t(x) - d_t(x).
\end{equation}

Subtracting the forward and backward marginal equations isolates the diffusion-induced discrepancy between the two drifts, yielding the following result.

\begin{proposition}{Osmotic Component}{d_component}
Under the canonical representative, the antisymmetric component is determined by the marginal score:
\begin{equation}
\label{eq:d_meaning}
  d_t(x)=\frac{\beta}{2}\nabla \log \pi_t(x).
\end{equation}
\begin{proof}
\textit{The result follows immediately from the definition of \(d_t\) in Eq.~\eqref{eq:d} together with the bidirectional consistency relation in Eq.~\eqref{eq:bidirectional_consistency}.}
\end{proof}
\end{proposition}

This shows that \(d_t\) is not an arbitrary degree of freedom: it is a density-dependent field induced by diffusion.
In particular, it encodes the effect of stochastic dynamics within a deterministic, density-dependent vector field.

In contrast, averaging the forward and backward marginal equations cancels the diffusion terms and isolates the symmetric component, yielding the following Proposition.
\begin{proposition}{Transport Component}{u_component}
The symmetric component \(u_t\) governs the same marginal evolution through the continuity equation:
\begin{equation}
\label{eq:u_continuity_eq}
  \partial_t \pi_t + \nabla \cdot \bigl(\pi_t u_t\bigr)=0.
\end{equation}
Equivalently, along deterministic characteristics generated by \(u_t\), with the Lagrangian derivative defined as 
\(\frac{D}{Dt} f(x,t) := \partial_t f(x,t) + u_t(x)\cdot \nabla f(x,t)\),
the log-density evolves according to
\begin{equation}
\label{eq:u_meaning}
  \frac{D}{Dt}\log \pi_t(x) = - \nabla \cdot u_t(x).
\end{equation}
\begin{proof}
\textit{From the definition of \(u_t\) in Eq.~\eqref{eq:u}, adding Equations~\ref{eq:forward_fpe} and~\ref{eq:backward_fpe} yields Eq.~\eqref{eq:u_continuity_eq}.
Since \(\pi_t>0\), we can divide Eq.~\eqref{eq:u_continuity_eq} by \(\pi_t\).
Using
\(\nabla \log \pi_t = \nabla \pi_t / \pi_t\) and
\(\partial_t \log \pi_t = \partial_t \pi_t / \pi_t\),
we obtain Eq.~\eqref{eq:u_meaning}.}
\end{proof}
\end{proposition}

The first identity, Eq.~\eqref{eq:u_continuity_eq}, shows that \(u_t\) serves as a deterministic transport field whose continuity equation reproduces the marginal evolution.
The second identity, Eq.~\eqref{eq:u_meaning}, reveals its local geometric meaning: along Lagrangian trajectories generated by \(u_t\), density increases in regions of compression, where \(\nabla\cdot u_t<0\), and decreases in regions of expansion, where \(\nabla\cdot u_t>0\).

\paragraph{Interpretation.}
The above results reveal the geometric roles of the two components.
The symmetric component \(u_t\) is the \textbf{transport field}: it carries probability mass and serves as the deterministic marginal dynamics of the stochastic process.
The antisymmetric component \(d_t\) is the \textbf{osmotic field}: it is determined by the marginal score \(\nabla \log \pi_t\) and captures diffusion-induced, density-dependent effects.
Importantly, the score does not represent sample-wise randomness itself; rather, it describes how diffusion affects the evolution of the marginal density.
Therefore, we call \(d_t\) osmotic because it is proportional to the score, analogous to the classical osmotic velocity in diffusion processes \citep{nelson1967dynamical}.

\paragraph{Connection to deterministic transport.}
This decomposition clarifies the relation between deterministic and stochastic generative dynamics.
In deterministic transport models, the learned velocity field directly serves as the marginal transport field, governing the density path through the continuity equation in Eq.~\eqref{eq:continuity_equation}.
In stochastic dynamics, by contrast, the forward drift contains both the transport component and the diffusion-induced score component, as reflected by the Fokker--Planck equation in Eq.~\eqref{eq:fokker_plank}.
Thus, deterministic and stochastic generative models share a common marginal transport backbone, but differ in how this backbone is realized at the level of sample dynamics: deterministic models directly parameterize \(u_t\), whereas stochastic models realize the forward drift as \(b_t^\rightarrow=u_t+d_t\).

%%%%%%%%%%%%%%%%%%%%%%%%%%%%%%%%%%%%%%%%%%%%%%%%%%%%%%%%%%%%%%%%%%
%%%%%%%%%%%%%%%%%%%%%%%%%%%%%%%%%%%%%%%%%%%%%%%%%%%%%%%%%%%%%%%%%%
\section{Bridge Matching for learning decomposed dynamics}
\label{sec:implementation}

To instantiate the proposed Transport--Osmotic decomposition in generative modeling, we introduce a flow-based training framework, which we refer to as \emph{Bridge Matching} (BM).
Given a base distribution \(\pi_0\) and a data distribution \(\pi_1\), the goal is to learn dynamics that transport samples along an intermediate probability path \(\{\pi_t\}_{t\in[0,1]}\).
Instead of parameterizing a single velocity field, BM learns two deterministic vector fields: a transport field \(u_\theta(x,t)\) and an osmotic field \(d_\phi(x,t)\).
The forward sampling dynamics are given by
\begin{equation}
\label{eq:bm_forward_dynamics}
    b_t^{\mathrm{BM}}(x)
    =
    u_\theta(x,t)+d_\phi(x,t),
\end{equation}
while the reverse direction can be represented by \(u_\theta(x,t)-d_\phi(x,t)\).
Thus, BM preserves the bidirectional structure of the decomposition while remaining implementable with deterministic neural vector fields.
Given target fields \(u^*(x,t)\) and \(d^*(x,t)\), we train the two components \(\lambda>0\) controls the training weight using a Flow Matching-style regression objective:
\begin{equation}
\label{eq:bm_loss}
\mathcal{L}_{\mathrm{BM}}
=
\mathbb{E}_{t,x_t}
\left[
\|u_\theta(x_t,t)-u^*(x_t,t)\|^2
+
\lambda
\|d_\phi(x_t,t)-d^*(x_t,t)\|^2
\right].
\end{equation}
% where \(\lambda>0\) controls the training weight of the osmotic component.

The main practical question is how to construct the targets \(u^*\) and \(d^*\).
Since \(d_t\) is tied to the marginal score \(\nabla \log \pi_t\), directly computing it is generally difficult when the intermediate density is unknown.
We therefore consider two constructions: a marginal formulation, which follows the distribution-level decomposition more directly, and a conditional formulation, which provides tractable closed-form targets.
See Appendix \ref{app:bm-formulations} for implementation details.  

\textbf{Marginal Bridge Matching (MBM)} defines the targets at the level of the intermediate marginal distribution.
Given samples \(x_t\sim\pi_t\) from a prescribed probability path, the transport target \(u_t^*\) is chosen to realize the marginal evolution through the continuity equation, whereas osmotic target is defined by the marginal score:
\begin{equation}
\label{eq:mbm_transport_continuity}
    \partial_t \pi_t + \nabla \cdot (\pi_t u_t^*) = 0, \quad d_t^*(x_t)=\frac{\beta_t}{2}\nabla_{x_t}\log \pi_t(x_t).
\end{equation}
This formulation follows the theoretical decomposition most directly, but requires access to the marginal score of the intermediate path.

\textbf{Conditional Bridge Matching (CBM)} avoids explicit marginal score estimation by defining a stochastic interpolation with tractable conditional density.
Let \(z\) denote an auxiliary conditioning variable, such as endpoint samples or latent noise, and define
\begin{equation}
\label{eq:conditional_path}
    x_t \sim p_t(x\mid z),
    \qquad 
    \pi_t(x)=\int p_t(x\mid z)p(z)\,dz.
\end{equation}
Given this conditional path, we construct conditional targets \(u_t^*(x_t,z)\) and \(d_t^*(x_t,z)\).
The transport target is taken as the velocity of the prescribed conditional evolution, while the osmotic target is given by the conditional score:
\begin{equation}
\label{eq:cbm_conditional_score}
    d_t^*(x_t,z)
    =
    \frac{\beta_t}{2}
    \nabla_{x_t}\log p_t(x_t\mid z).
\end{equation}
As in Conditional Flow Matching \citep{lipman2023flow, tong2023minibatchot}, training on these conditional targets yields marginal vector fields in expectation, with the conditional score recovering the marginal score through \(\mathbb{E}[\nabla_x \log p_t(x\mid z)\mid x_t=x]=\nabla_x\log\pi_t(x)\).

\begin{figure}[t]
    \centering
    \includegraphics[width=0.95\linewidth,clip,trim={25pt 25pt 5pt 15pt}]{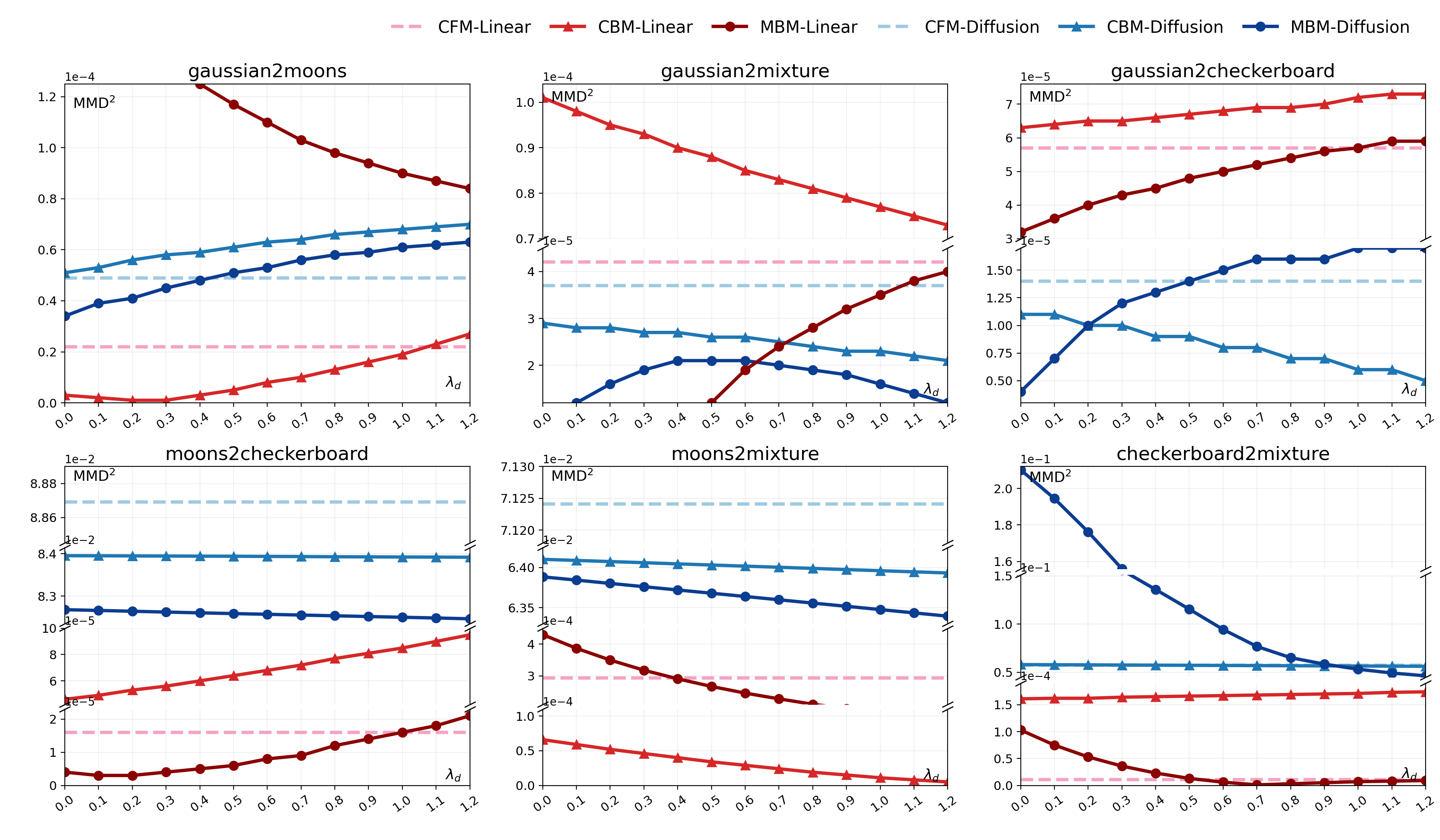}
    \vspace{0pt}
    \caption{
    MMD\(^2\) (lower is better) between generated and target samples on 2D transport tasks, for different source--target pairs.
    Horizontal axis is the recombination weight \(\lambda_d\).
    % Dashed lighter curves denote CFM baselines, while solid darker curves denote BM variants.
    }
    \label{fig:2d_quantitative_comparison}
    \vspace{-12pt}
\end{figure}

%%%%%%%%%%%%%%%%%%%%%%%%%%%%%%%%%%%%%%%%%%%%%%%%%%%%%%%%%%%%%%%%%%
%%%%%%%%%%%%%%%%%%%%%%%%%%%%%%%%%%%%%%%%%%%%%%%%%%%%%%%%%%%%%%%%%%
\section{Numerical examples}
\label{sec:experiments}
% We evaluate the BM framework from two complementary perspectives.
% We first assess generative performance on 2D and image datasets by comparing BM with FM baselines under different probability paths and sampling-time recombination weights, \(b_t=\lambda_u u_\theta+\lambda_d d_\phi\).
% We then analyze the learned transport and osmotic components to demonstrate the controllability and interpretability induced by the decomposition.
We use these numerical examples as controlled analyses of the Transport--Osmotic decomposition, focusing on how recombining the learned fields as
\(b_t = u_t + \lambda_d d_t\)
affects the resulting generative dynamics.
See Appendix \ref{app:2d-generation-details} and \ref{app:image-generation-details} for implementation details, and Appendix \ref{app:additional-results} for additional results. 
%%%%%%%%%%%%%%%%%%%%%%%%%%%%%%%%%%%%%%%%%%%%%%%%%%%%%%%%%%%%%%%%%%
% \subsection{Generative performance and ablation studies under different probability paths}
% \label{subsec:generation_performance}

\vspace{-6pt}
\paragraph{2D generation.}
We first evaluate BM on 2D transport tasks involving different source--target pairs, including checkerboard, moons, and Gaussian mixture distributions.
We compare six targets: CFM-Linear, CFM-Diffusion, CBM-Linear, CBM-Diffusion, MBM-Linear, and MBM-Diffusion.
Here, CFM denotes Conditional Flow Matching with either the standard linear path or a diffusion path as baseline. 
For all BM variants, samples are generated using the recombined dynamics with various osmotic weight \(\lambda_d\).
Although the standard linear CFM path is deterministic, we additionally construct tube-like linear paths for the BM variants (CBM-Linear and MBM-Linear).
This allows us to compare against the deterministic linear CFM baseline while evaluating whether an osmotic component can improve endpoint matching (see Appendix~\ref{app:target-details}).
% within an otherwise linear transport geometry 
For diffusion-based paths, the comparison evaluates whether explicitly decomposing the dynamics into transport and osmotic components improves generation and provides additional control over the generated samples.

\vspace{-2pt}
% NOTE: The paragraph is too long; split here to show clearly where the result starts
As shown in Fig.~\ref{fig:2d_quantitative_comparison}, BM targets achieve competitive endpoint matching compared with FM baselines across different transport tasks.
The effect of \(\lambda_d\) is strongly task- and path-dependent; for some source-target pairs, especially those involving more irregular or multimodal target structures (mixture), increasing the osmotic contribution (\(d_t \uparrow\)) improves the endpoint match (MMD\(^2 \downarrow\)).
For other cases (moons), however, an overly large \(d_t\) can degrade performance.
These results suggest that the useful strength of \(d_t\) depends on the geometry of the probability path and the multimodal structure of the endpoint distributions, which together determine the balance between coherent transport through \(u_t\) and density-dependent reshaping through \(d_t\).
Therefore, \(d_t\) should not be interpreted as a simple additive correction that always improves generation quality.
Instead, it provides a controllable component whose optimal contribution depends on the underlying transport problem.

\vspace{-6pt}
\paragraph{Image generation.}
For image generation tasks, we evaluate on CIFAR-10, ImageNet-32, and ImageNet-64 \citep{krizhevsky2009learning,deng2009imagenet,chrabaszcz2017downsampled}.
We focus on diffusion-path CFM and the corresponding CBM model, as this setting directly evaluates the effect of decomposing stochastic-path dynamics while allowing a direct comparison with the CFM baseline.
Table~\ref{tab:image_generation_lambda_ablation} shows three main observations.
First, CBM achieves comparable or better generation quality than the CFM baseline, while additionally providing an explicit decomposition of the sampling dynamics.
Second, the optimal recombination weight \(\lambda_d\) depends on the evaluation metric: the best FID, KID, and IS are not necessarily achieved by the same value of \(\lambda_d\), suggesting that adjusting the osmotic contribution provides a practical mechanism for trading off different aspects of generation quality.
Third, the best performance is not always obtained at \(\lambda_d=1\), even though this corresponds to the direct reconstruction of the forward drift under the decomposition in Eq.~\eqref{eq:bm_forward_dynamics}.
This indicates that the theoretically defined forward dynamics are not necessarily optimal for finite-capacity neural models, finite-step solvers, or a given evaluation metric, and that sampling-time recombination provides an additional useful degree of freedom.
\begin{table}[t]
\centering
\vspace{0pt}
\caption{
Generation performance and recombination ablation on image datasets.
This table compares diffusion-path CFM with the corresponding CBM model.
For CBM, different recombination weights \(\lambda_d\) are evaluated at sampling time using the same trained model.
}
\vspace{2pt}
\label{tab:image_generation_lambda_ablation}
\renewcommand{\arraystretch}{0.95}
\resizebox{0.7\linewidth}{!}{
\begin{tabular}{llcccc}
\toprule
Dataset & Method & \(\lambda_d\) & FID \(\downarrow\) & KID \((\times 10^{-3})\downarrow\) & IS \(\uparrow\) \\
\midrule
\multirow{7}{*}{CIFAR-10} 
& CFM                  & \(\text{--}\) & \(5.21\) & \(2.81 \pm 0.65\) & \(8.80 \pm 0.10\) \\
\cmidrule(lr){2-6}
& \multirow{6}{*}{CBM} & \(0.00\)      & \(9.93\) & \(6.83 \pm 0.71\) & \(\mathbf{9.29 \pm 0.12}\) \\
&                      & \(0.25\)      & \(7.39\) & \(4.04 \pm 0.52\) & \(9.19 \pm 0.11\) \\
&                      & \(0.50\)      & \(5.77\) & \(\mathbf{2.52 \pm 0.43}\) & \(9.05 \pm 0.10\) \\
&                      & \(0.75\)      & \(\mathbf{5.34}\) & \(2.53 \pm 0.49\) & \(8.89 \pm 0.09\) \\
&                      & \(1.00\)      & \(5.78\) & \(3.59 \pm 0.64\) & \(8.73 \pm 0.07\) \\
&                      & \(1.50\)      & \(7.71\) & \(6.64 \pm 0.87\) & \(8.53 \pm 0.08\) \\
\midrule
\multirow{7}{*}{ImageNet-32}
& CFM                  & \(\text{--}\) & \(13.11\) & \(2.65 \pm 0.36\) & \(10.32 \pm 0.16\) \\
\cmidrule(lr){2-6}
& \multirow{6}{*}{CBM} & \(0.00\)      & \(15.13\) & \(4.20 \pm 0.23\) & \(\mathbf{11.15 \pm 0.37}\) \\
&                      & \(0.25\)      & \(13.88\) & \(3.10 \pm 0.25\) & \(11.03 \pm 0.38\) \\
&                      & \(0.50\)      & \(13.07\) & \(2.50 \pm 0.29\) & \(10.93 \pm 0.40\) \\
&                      & \(0.75\)      & \(\mathbf{12.75}\) & \(\mathbf{2.43 \pm 0.39}\) & \(10.78 \pm 0.43\) \\
&                      & \(1.00\)      & \(12.81\) & \(2.75 \pm 0.47\) & \(10.64 \pm 0.46\) \\
&                      & \(1.50\)      & \(13.52\) & \(3.87 \pm 0.62\) & \(10.44 \pm 0.45\) \\
\midrule
\multirow{7}{*}{ImageNet-64} 
& CFM                  & \(\text{--}\) & \(13.39\) & \(2.20 \pm 0.37\) & \(11.80 \pm 0.19\) \\
\cmidrule(lr){2-6}
& \multirow{6}{*}{CBM} & \(0.00\)      & \(16.12\) & \(3.59 \pm 0.40\) & \(\mathbf{12.37 \pm 0.27}\) \\
&                      & \(0.25\)      & \(14.78\) & \(2.72 \pm 0.39\) & \(12.26 \pm 0.26\) \\
&                      & \(0.50\)      & \(13.81\) & \(2.12 \pm 0.37\) & \(12.13 \pm 0.24\) \\
&                      & \(0.75\)      & \(13.40\) & \(\mathbf{2.00 \pm 0.36}\) & \(11.99 \pm 0.26\) \\
&                      & \(1.00\)      & \(\mathbf{13.30}\) & \(2.26 \pm 0.24\) & \(11.84 \pm 0.26\) \\
&                      & \(1.50\)      & \(14.26\) & \(3.26 \pm 0.21\) & \(11.65 \pm 0.26\) \\
\bottomrule
\end{tabular}
}
\vspace{-6pt}
\end{table}

\begin{figure}[t]
    \centering
    \includegraphics[width=0.8\linewidth]{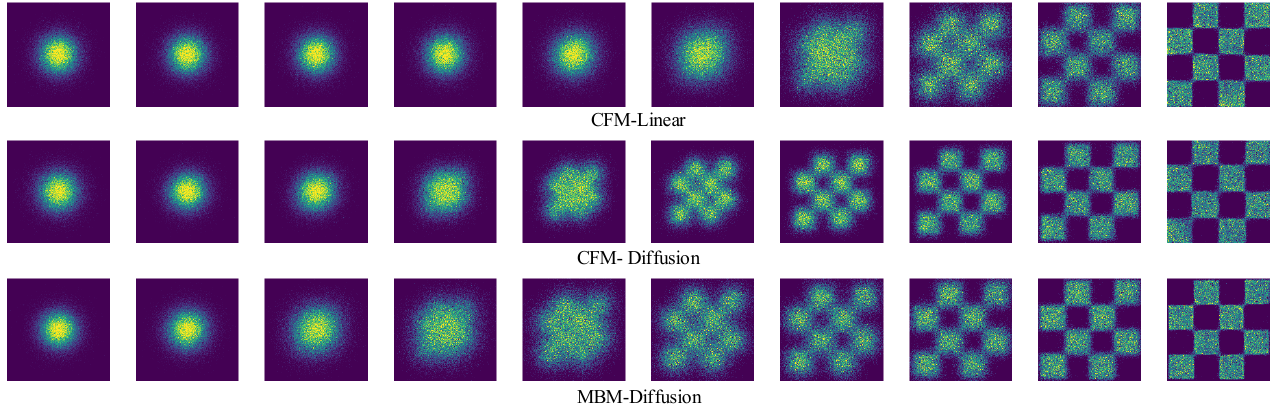}
    \vspace{-3pt}
    \caption{
    Marginal evolution comparison on a 2D Gaussian-to-checkerboard task.
    This visualization includes intermediate generated distributions for CFM-Linear, CFM-Diffusion, and MBM-Diffusion.
    }
    \label{fig:2d_marginal_evolution}
    \vspace{1.5ex}
    \includegraphics[width=0.85\linewidth]{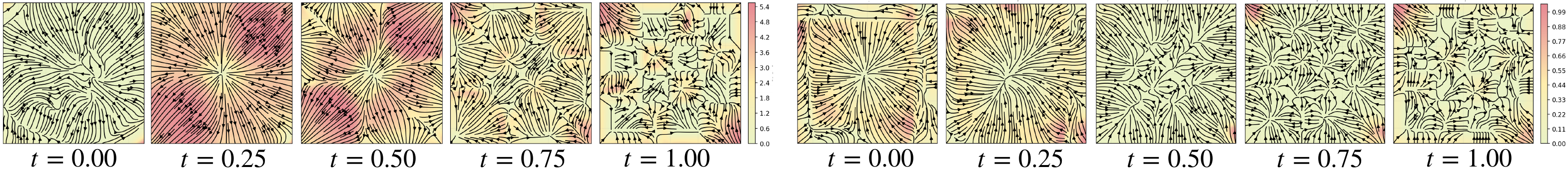}
    \caption{
    Field-level visualization of learned decomposition: transport \(u_t\) (\emph{left}) and osmotic \(d_t\) (\emph{right}).
    }
    \label{fig:field_visualization}
\vspace{-8pt}
\end{figure}

\vspace{-6pt}
\paragraph{2D marginal evolution.}
To compare trajectory-level behavior, we visualize intermediate marginals on the Gaussian-to-checkerboard task.
As shown in Fig.~\ref{fig:2d_marginal_evolution}, the linear FM path transports samples through a relatively direct interpolation and gradually forms the checkerboard structure near the end of the trajectory.
In contrast, diffusion-path FM introduces a more dispersed intermediate evolution before concentrating into the target modes.
MBM-Diffusion induces earlier marginal reshaping, producing diffusion-like spreading before progressively concentrating samples into the checkerboard structure.
% through the osmotic component
These results suggest that the decomposition affects not only endpoint matching, but also the intermediate redistribution of probability mass along the sampling trajectory.

\vspace{-6pt}
\paragraph{Field-level interpretation.}
We visualize the vector fields induced by \(u_t\) and \(d_t\) on the Gaussian-to-Checkerboard task.
As shown in Fig.~\ref{fig:field_visualization}, \(u_t\) forms coherent large-scale motion that moves probability mass across the space.
In contrast, \(d_t\) has a smaller magnitude but exhibits structured spatial patterns, especially at early and intermediate times, indicating its role in locally reshaping the evolving density rather than dominating the global transport.
This supports the interpretation that \(u_t\) provides the main marginal transport backbone, while \(d_t\) acts as a density-dependent correction field that adjusts the sample dynamics along the probability path.

%%%%%%%%%%%%%%%%%%%%%%%%%%%%%%%%%%%%%%%%%%%%%%%%%%%%%%%%%%%%%%%%%%

%%%%%%%%%%%%%%%%%%%%%%%%%%%%%%%%%%%%%%%%%%%%%%%%%%%%%%%%%%%%%%%%%%

%%%%%%%%%%%%%%%%%%%%%%%%%%%%%%%%%%%%%%%%%%%%%%%%%%%%%%%%%%%%%%%%%%

%%%%%%%%%%%%%%%%%%%%%%%%%%%%%%%%%%%%%%%%%%%%%%%%%%%%%%%%%%%%%%%%%%
% \vspace{-5pt}
% \begin{wrapfigure}[10]{r}{0.49\linewidth}
% \vspace{-12pt}
% \centering
% \includegraphics[width=0.95\linewidth]{song_bm_nips26/fig/res/2_image_lambda_visualization.pdf}
% \caption{
% Sampling-time controllability through recombination: \(\lambda_d=0\) tends to preserve sharper local structures with more high-frequency details, whereas \(\lambda_d=1\) produces smoother but slightly more blurred samples.
% }
% \label{fig:image_lambda_visualization}
% \vspace{-8pt}
% \end{wrapfigure}

% NOTE: lining up two wrapfigs vertically also looks a bit unfamiliar...

\begin{figure}[t]
\begin{minipage}[t]{0.54\linewidth}
    \vspace*{0pt}
    \centering
    \includegraphics[width=0.95\linewidth]{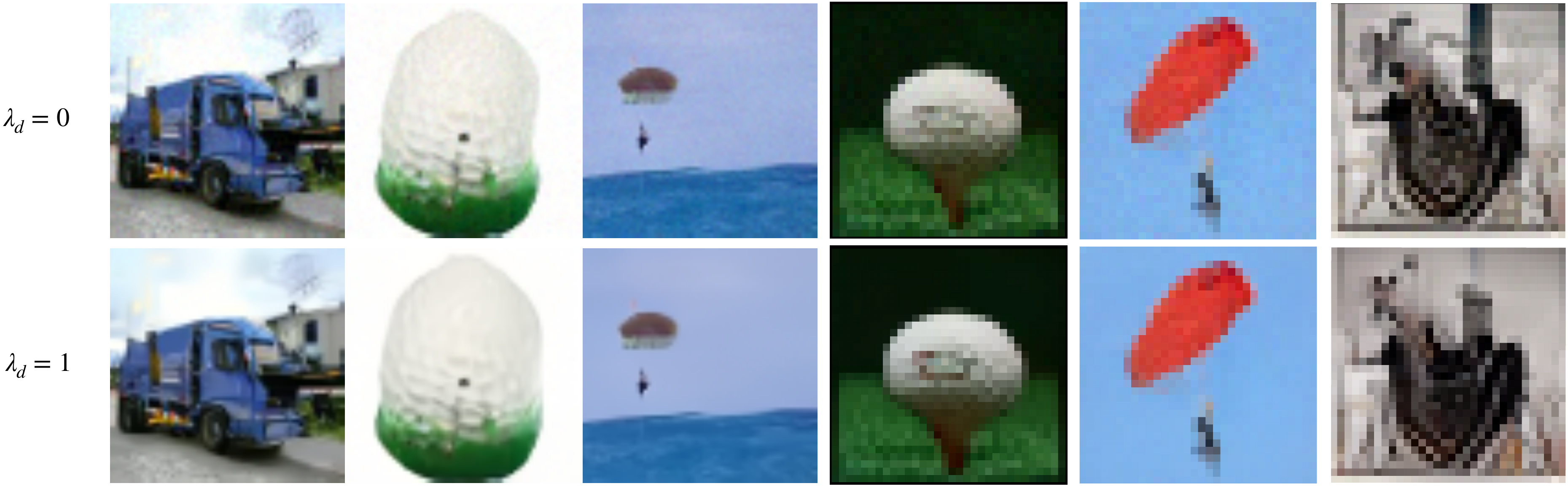}
    \caption{
    Sampling-time controllability through recombination: \(\lambda_d=0\) (\emph{upper}) tends to preserve sharper, high-frequency local structures, and \(\lambda_d=1\) (\emph{lower}) produces smoother but slightly more blurred samples.
    }
    \label{fig:image_lambda_visualization}
\end{minipage}
\hfill
\begin{minipage}[t]{0.42\linewidth}
    \vspace*{0pt}
    \centering
    \includegraphics[width=0.80\linewidth]{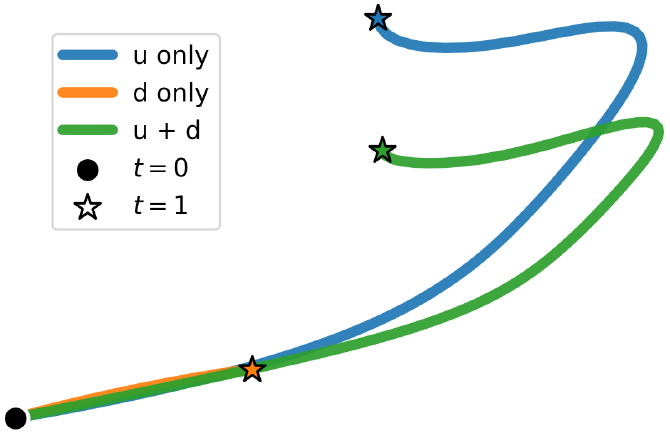}
    \caption{
    Particle-level comparison under decomposed fields \(u_t\) and \(d_t\).
    }
    \label{fig:single_particle_trajectory}
\end{minipage}
\vspace{-1.5ex}
\end{figure}

\paragraph{Sampling-time controllability.}
Fig.~\ref{fig:image_lambda_visualization} illustrates the sampling-time controllability enabled by the proposed decomposition.
By varying the osmotic weight \(\lambda_d\) while keeping the learned model fixed, the generated samples exhibit different visual characteristics.
It suggests that \(\lambda_d\) provides a practical mechanism for adjusting image appearance according to different user preferences, without retraining the model, thereby exposing a controllable trade-off between sharpness and smoothness.

%%%%%%%%%%%%%%%%%%%%%%%%%%%%%%%%%%%%%%%%%%%%%%%%%%%%%%

% \begin{wrapfigure}[11]{r}{0.48\linewidth}
% \vspace{-10pt}
% \centering
% \includegraphics[width=0.80\linewidth]{fig/res/2_single_particle_trajectory.pdf}
% \caption{
% Particle-level comparison under decomposed fields \(u_t\) and \(d_t\).
% }
% \label{fig:single_particle_trajectory}
% \vspace{-8pt}
% \end{wrapfigure}
% \vspace{-8pt}
%
\paragraph{Particle-level interpretation.}
We further illustrate the distinct roles of the two components by integrating the same initial particle under \(u_t\), \(d_t\), and \(u_t+d_t\), as shown in Fig.~\ref{fig:single_particle_trajectory}.
The \(u_t\)-only trajectory reflects coherent transport, whereas the \(d_t\)-only trajectory follows a distinct density-dependent correction direction.
The recombined trajectory largely preserves the transport trend induced by \(u_t\), while being shifted by the osmotic contribution from \(d_t\).
This particle-level behavior provides an intuitive view of how the two components interact during generation.

\WFclear

%%%%%%%%%%%%%%%%%%%%%%%%%%%%%%%%%%%%%%%%%%%%%%%%%%%%%%%%%%%%%%%%%%
%%%%%%%%%%%%%%%%%%%%%%%%%%%%%%%%%%%%%%%%%%%%%%%%%%%%%%%%%%%%%%%%%%
\vspace{-6pt}
\section{Discussion}
\label{sec:related_work}
\vspace{-6pt}
\paragraph{Relation to prior works.} Modern generative models can be viewed through the lens of probability transport.
Normalizing flows model distribution transformation through invertible maps \citep{dinh2017realnvp}, while diffusion models learn stochastic noising and denoising processes \citep{ho2020ddpm,karras2022edm}.
Flow Matching instead formulate generation through deterministic velocity fields along probability paths \citep{lipman2023flow}, with recent variants improving coupling, path geometry, sampling efficiency, and one-step or bidirectional generation \citep{lee2023curvature,improveRF2024,geng2025meanflow,lu2025biflow}.
Scalable image generators further combine diffusion or flow objectives with latent representations and transformer architectures \citep{rombach2022latent,peebles2023scalable,ma2024sit}.
Optimal Transport and entropic regularization provide a principled foundation for deterministic and stochastic distribution transformations \citep{villani2009optimal,leonard2014survey}, while Schr{\"o}dinger Bridge formulates stochastic optimal transport through forward and backward dynamics linked by marginal scores \citep{haussmann1986reverse,shi2023dsbm,tong2024simulationfree,liu2024gsbm}.
Recent bridge-based methods extend this view to conditional generation, stable sampling, pretrained diffusion initialization, and non-memoryless or adjoint formulations \citep{gottwald2024stable,huang2025conditionalsb,tang2025pretrainedsb,shin2026asbm}.
Although these perspectives connect flows, diffusion models, and bridge-based methods, they often absorb transport and stochastic density-shaping effects into a single dynamic field.
Our work complements prior studies on scores, stochastic interpolants, trajectory straightening, and path regularization \citep{liu2023rectified,lee2023curvature} by explicitly decomposing generative dynamics into transport and osmotic components.

\vspace{-6pt}
\paragraph{Conclusion, limitations and future work.}
We proposed a Transport--Osmotic decomposition of stochastic generative dynamics into a marginal transport field \(u_t\) and an osmotic field \(d_t\), separating two components often hidden within a single effective drift. 
Bridge Matching learns these fields and recombines them at sampling time, enabling controllable adjustment of endpoint quality and intermediate probability paths across 2D and image-generation experiments. 
This provides not only an unified perspective on deterministic and stochastic dynamics, but an interpretable perspective on how these dynamics shape probability transport.  
The main limitations are that, the decomposition applies most directly to field-based generative dynamics represented by fields, and learning separate fields introduces additional modeling and computational overhead. 
Reducing this overhead while preserving interpretability and controllability in downstream applications is an important direction for future work.
In addition, the bidirectional structure preserved by the decomposition may support interactive sampling procedures, allowing generation to be paused, reversed, and regenerated from intermediate states, which could provide a basis for real-time editing in generative models.

%%%%%%%%%%%%%%%%%%%%%%%%%%%%%%%%%%%%%%%%%%%%%%%%%%%%%%%%%%%%%%%%%%
%%%%%%%%%%%%%%%%%%%%%%%%%%%%%%%%%%%%%%%%%%%%%%%%%%%%%%%%%%%%%%%%%%
\section{Acknowledgment}
This work was supported by JST PRESTO JPMJPR24T6, JSPS KAKENHI JP25H01454 and JP26K02968.
This work was also supported by JST SPRING, Grant Number JPMJSP2108. 
We are grateful to Henrik Krauss and Omar Sharif for their valuable comments and help with revising the paper.
%%%%%%%%%%%%%%%%%%%%%%%%%%%%%%%%%%%%%%%%%%%%%%%%%%%%%%%%%%%%%%%%%%
%%%%%%%%%%%%%%%%%%%%%%%%%%%%%%%%%%%%%%%%%%%%%%%%%%%%%%%%%%%%%%%%%%

\newpage
\bibliographystyle{plainnat}
\bibliography{ref}
%%%%%%%%%%%%%%%%%%%%%%%%%%%%%%%%%%%%%%%%%%%%%%%%%%%%%%%%%%%%%%%%%%
%%%%%%%%%%%%%%%%%%%%%%%%%%%%%%%%%%%%%%%%%%%%%%%%%%%%%%%%%%%%%%%%%%
\newpage
\input{appendix}
%%%%%%%%%%%%%%%%%%%%%%%%%%%%%%%%%%%%%%%%%%%%%%%%%%%%%%%%%%%%%%%%%%
%%%%%%%%%%%%%%%%%%%%%%%%%%%%%%%%%%%%%%%%%%%%%%%%%%%%%%%%%%%%%%%%%%
% \clearpage
% \input{checklist.tex}
%%%%%%%%%%%%%%%%%%%%%%%%%%%%%%%%%%%%%%%%%%%%%%%%%%%%%%%%%%%%%%%%%%
%%%%%%%%%%%%%%%%%%%%%%%%%%%%%%%%%%%%%%%%%%%%%%%%%%%%%%%%%%%%%%%%%%

\end{document}

%% file: appendix.tex
\appendix
%%%%%%%%%%%%%%%%%%%%%%%%%%%%%%%%%%%%%%%%%%%%%%%%%%%%%%%%%%%%%%%%%%
%%%%%%%%%%%%%%%%%%%%%%%%%%%%%%%%%%%%%%%%%%%%%%%%%%%%%%%%%%%%%%%%%%
\section{Derivations and proofs}
\label{app:derivations}

%%%%%%%%%%%%%%%%%%%%%%%%%%%%%%%%%%%%%%%%%%%%%%%%%%%%%%%%%%%%%%%%%%
\subsection{Deterministic transport and flow maps}
\label{app:deterministic-transport}

We briefly recall how a time-dependent velocity field induces a deterministic probability transport path.
Let \(v_t:\mathbb{R}^d\to\mathbb{R}^d\) be a sufficiently regular velocity field.
For an initial point \(x_0\), define the flow map \(\phi_t\) as the solution of the ordinary differential equation
\begin{equation}
    \frac{d}{dt}\phi_t(x_0) = v_t(\phi_t(x_0)),
    \qquad
    \phi_0(x_0)=x_0 .
\end{equation}
If \(X_0\sim \pi_0\), then the deterministic process generated by this flow is
\begin{equation}
    X_t=\phi_t(X_0).
\end{equation}
Therefore, for any measurable set \(A\subseteq\mathbb{R}^d\),
\begin{equation}
    \mathbb{P}(X_t\in A)
    =
    \mathbb{P}(\phi_t(X_0)\in A)
    =
    \mathbb{P}(X_0\in \phi_t^{-1}(A)).
\end{equation}
Using \(X_0\sim \pi_0\), this becomes
\begin{equation}
    \mathbb{P}(X_t\in A)
    =
    \int_{\phi_t^{-1}(A)} \pi_0(x)\,dx.
\end{equation}
This coincides with the definition of the pushforward distribution, and hence the marginal law of \(X_t\) is
\begin{equation}
    \pi_t = [\phi_t]_{\#}\pi_0 .
\end{equation}
Equivalently, the sample-level dynamics can be written directly as
\begin{equation}
    \frac{dX_t}{dt}=v_t(X_t),
\end{equation}
which is Eq.~\eqref{eq:ode_for_fm} in the main text.

%%%%%%%%%%%%%%%%%%%%%%%%%%%%%%%%%%%%%%%%%%%%%%%%%%%%%%%%%%%%%%%%%%
\subsection{Derivation of the continuity equation}
\label{app:continuity-equation}

We derive the continuity equation associated with the deterministic dynamics in Eq.~\eqref{eq:ode_for_fm}.
Let \(X_t\) evolve according to the ODE
\begin{equation}
    \frac{dX_t}{dt}=v_t(X_t),
\end{equation}
and let \(\pi_t\) denote the density of \(X_t\).
For any smooth test function \(f:\mathbb{R}^d\to\mathbb{R}\) with compact support, we have
\begin{equation}
    \frac{d}{dt}\mathbb{E}_{X_t\sim \pi_t}[f(X_t)]
    =
    \frac{d}{dt}\int f(x)\pi_t(x)\,dx .
\end{equation}
On the one hand, differentiating the integral gives
\begin{equation}
    \frac{d}{dt}\int f(x)\pi_t(x)\,dx
    =
    \int f(x)\partial_t\pi_t(x)\,dx .
\end{equation}
On the other hand, using the chain rule along trajectories of the ODE,
\begin{equation}
    \frac{d}{dt}f(X_t)
    =
    \nabla f(X_t)^\top v_t(X_t).
\end{equation}
Taking expectation gives
\begin{equation}
    \frac{d}{dt}\mathbb{E}_{X_t\sim \pi_t}[f(X_t)]
    =
    \int \nabla f(x)^\top v_t(x)\pi_t(x)\,dx .
\end{equation}
Using integration by parts and assuming boundary terms vanish, we obtain
\begin{equation}
    \int \nabla f(x)^\top v_t(x)\pi_t(x)\,dx
    =
    -\int f(x)\nabla\cdot\bigl(\pi_t(x)v_t(x)\bigr)\,dx .
\end{equation}
Therefore,
\begin{equation}
    \int f(x)
    \left[
    \partial_t\pi_t(x)
    +
    \nabla\cdot\bigl(\pi_t(x)v_t(x)\bigr)
    \right]dx
    =
    0 .
\end{equation}
Since this identity holds for arbitrary smooth test functions \(f\), we conclude that
\begin{equation}
    \partial_t\pi_t(x)
    +
    \nabla\cdot\bigl(\pi_t(x)v_t(x)\bigr)
    =
    0,
\end{equation}
which is Eq.~\eqref{eq:continuity_equation} in the main text.

This equation states that the density changes only through the flux \(\pi_t v_t\) induced by the deterministic velocity field.
Thus, while Eq.~\eqref{eq:ode_for_fm} describes individual sample trajectories, Eq.~\eqref{eq:continuity_equation} describes the corresponding evolution of the full probability density.

%%%%%%%%%%%%%%%%%%%%%%%%%%%%%%%%%%%%%%%%%%%%%%%%%%%%%%%%%%%%%%%%%%
\subsection{Stochastic transport dynamics}
\label{app:stochastic-transport}

We next recall the stochastic counterpart of deterministic transport.
Instead of evolving samples through a deterministic ODE, stochastic transport evolves samples through an It\^{o} stochastic differential equation of the form
\begin{equation}
    dX_t = b_t(X_t)\,dt + \sqrt{\beta_t}\,dW_t ,
\end{equation}
where \(b_t:\mathbb{R}^d\to\mathbb{R}^d\) is the drift field, \(\beta_t>0\) is a time-dependent scalar diffusion coefficient, and \(W_t\) denotes standard Brownian motion.

Equivalently, this SDE can be written in integral form as
\begin{equation}
    X_t
    =
    X_0
    +
    \int_0^t b_s(X_s)\,ds
    +
    \int_0^t \sqrt{\beta_s}\,dW_s .
\end{equation}
The first integral represents the accumulated deterministic drift, while the second It\^{o} integral represents the accumulated stochastic perturbation from Brownian motion.
Therefore, unlike the deterministic dynamics
\begin{equation}
    \frac{dX_t}{dt}=v_t(X_t),
\end{equation}
the state \(X_t\) is not determined solely by \(X_0\), but also by the Brownian path \(\{W_s\}_{s\leq t}\).

As a result, the marginal density \(\pi_t\) of \(X_t\) is induced by both the directional drift \(b_t\) and the diffusion coefficient \(\beta_t\).
This stochastic process therefore realizes a probability path \(\{\pi_t\}_{t\in[0,1]}\) through sample-level stochastic dynamics, which is Eq.~\eqref{eq:sde_for_diffusion} in the main text.

%%%%%%%%%%%%%%%%%%%%%%%%%%%%%%%%%%%%%%%%%%%%%%%%%%%%%%%%%%%%%%%%%%
\subsection{Derivation of the Fokker--Planck equation}
\label{app:fokker-planck}

We derive the Fokker--Planck equation associated with the stochastic dynamics in Eq.~\eqref{eq:sde_for_diffusion}.
Consider the It\^{o} SDE
\begin{equation}
    dX_t = b_t(X_t)\,dt + \sqrt{\beta_t}\,dW_t ,
\end{equation}
where \(W_t\) is standard Brownian motion and \(\beta_t>0\) is a time-dependent scalar diffusion coefficient.
Let \(\pi_t\) denote the density of \(X_t\).

For any smooth test function \(f:\mathbb{R}^d\to\mathbb{R}\) with compact support, It\^{o}'s formula gives
\begin{equation}
    df(X_t)
    =
    \nabla f(X_t)^\top b_t(X_t)\,dt
    +
    \frac{\beta_t}{2}\Delta f(X_t)\,dt
    +
    \sqrt{\beta_t}\nabla f(X_t)^\top dW_t .
\end{equation}
Taking expectation, the stochastic integral has zero mean, so
\begin{equation}
    \frac{d}{dt}\mathbb{E}_{X_t\sim \pi_t}[f(X_t)]
    =
    \mathbb{E}_{X_t\sim \pi_t}
    \left[
    \nabla f(X_t)^\top b_t(X_t)
    +
    \frac{\beta_t}{2}\Delta f(X_t)
    \right].
\end{equation}
Writing the expectation in density form gives
\begin{equation}
    \frac{d}{dt}\int f(x)\pi_t(x)\,dx
    =
    \int
    \left[
    \nabla f(x)^\top b_t(x)
    +
    \frac{\beta_t}{2}\Delta f(x)
    \right]
    \pi_t(x)\,dx .
\end{equation}
The left-hand side is
\begin{equation}
    \frac{d}{dt}\int f(x)\pi_t(x)\,dx
    =
    \int f(x)\partial_t\pi_t(x)\,dx .
\end{equation}
For the drift term, integration by parts gives
\begin{equation}
    \int \nabla f(x)^\top b_t(x)\pi_t(x)\,dx
    =
    -\int f(x)\nabla\cdot\bigl(\pi_t(x)b_t(x)\bigr)\,dx .
\end{equation}
For the diffusion term, applying integration by parts twice gives
\begin{equation}
    \int \Delta f(x)\pi_t(x)\,dx
    =
    \int f(x)\Delta \pi_t(x)\,dx ,
\end{equation}
where boundary terms vanish due to the compact support of \(f\) or suitable decay conditions.

Combining these identities, we obtain
\begin{equation}
    \int f(x)\partial_t\pi_t(x)\,dx
    =
    \int f(x)
    \left[
    -\nabla\cdot\bigl(\pi_t(x)b_t(x)\bigr)
    +
    \frac{\beta_t}{2}\Delta \pi_t(x)
    \right]dx .
\end{equation}
Since this holds for arbitrary smooth test functions \(f\), we conclude that
\begin{equation}
    \partial_t \pi_t(x)
    =
    -\nabla\cdot\bigl(\pi_t(x)b_t(x)\bigr)
    +
    \frac{\beta_t}{2}\Delta \pi_t(x),
\end{equation}
which is Eq.~\eqref{eq:fokker_plank} in the main text.

The Fokker--Planck equation is the stochastic analogue of the continuity equation.
The first term,
\(-\nabla\cdot(\pi_t b_t)\), describes density change due to directional drift, while the second term,
\(\frac{\beta_t}{2}\Delta\pi_t\), describes density spreading due to diffusion.
When \(\beta_t=0\), the diffusion term disappears and the equation reduces to the deterministic continuity equation with velocity field \(b_t\).

%%%%%%%%%%%%%%%%%%%%%%%%%%%%%%%%%%%%%%%%%%%%%%%%%%%%%%%%%%%%%%%%%%
\subsection{Derivation of the forward--backward drift relation}
\label{app:forward-backward-drift}

We derive the relation between the forward and backward drifts of a diffusion process with shared marginal density path.
For simplicity, consider a diffusion process with constant scalar diffusion coefficient \(\beta>0\), forward drift \(b_t^\rightarrow\), and marginal density \(\pi_t\).
The forward-time dynamics are
\begin{equation}
    dX_t = b_t^\rightarrow(X_t)\,dt + \sqrt{\beta}\,dW_t .
\end{equation}
The corresponding forward Fokker--Planck equation is
\begin{equation}
\label{eq:app_forward_fpe}
    \partial_t \pi_t(x)
    =
    -\nabla\cdot\bigl(\pi_t(x)b_t^\rightarrow(x)\bigr)
    +
    \frac{\beta}{2}\Delta \pi_t(x).
\end{equation}

The same diffusion process can also be represented backward in time.
Let \(\tau=1-t\) and define the reverse-time process \(Y_\tau=X_{1-\tau}\).
Its reverse-time dynamics can be written as
\begin{equation}
    dY_\tau
    =
    \tilde b_\tau(Y_\tau)\,d\tau
    +
    \sqrt{\beta}\,d\bar W_\tau ,
\end{equation}
where \(\tilde b_\tau\) is the reverse-time drift and \(\bar W_\tau\) denotes Brownian motion with respect to the reversed filtration.
The marginal density of \(Y_\tau\) is
\begin{equation}
    \tilde\pi_\tau(x)=\pi_{1-\tau}(x).
\end{equation}
The standard Fokker--Planck equation in the reverse-time variable \(\tau\) is therefore
\begin{equation}
\label{eq:app_reverse_fpe_tau}
    \partial_\tau \tilde\pi_\tau(x)
    =
    -\nabla\cdot\bigl(\tilde\pi_\tau(x)\tilde b_\tau(x)\bigr)
    +
    \frac{\beta}{2}\Delta \tilde\pi_\tau(x).
\end{equation}
Since \(\tilde\pi_\tau(x)=\pi_{1-\tau}(x)\), we have
\begin{equation}
    \partial_\tau \tilde\pi_\tau(x)
    =
    -\partial_t \pi_t(x),
    \qquad t=1-\tau .
\end{equation}
Throughout this work, we express the backward drift in the original forward-time orientation by defining
\begin{equation}
    b_t^\leftarrow(x):=-\tilde b_{1-t}(x).
\end{equation}
Thus, \(b_t^\leftarrow\) is not the raw reverse-time drift, but the backward drift reoriented in the original forward-time coordinate.
Substituting these identities into Eq.~\eqref{eq:app_reverse_fpe_tau} gives
\begin{equation}
    -\partial_t\pi_t(x)
    =
    -\nabla\cdot\bigl(\pi_t(x)\tilde b_{1-t}(x)\bigr)
    +
    \frac{\beta}{2}\Delta\pi_t(x).
\end{equation}
Using \(\tilde b_{1-t}(x)=-b_t^\leftarrow(x)\), we obtain
\begin{equation}
    -\partial_t\pi_t(x)
    =
    \nabla\cdot\bigl(\pi_t(x)b_t^\leftarrow(x)\bigr)
    +
    \frac{\beta}{2}\Delta\pi_t(x).
\end{equation}
Multiplying both sides by \(-1\), the backward marginal equation expressed in the original time parameter becomes
\begin{equation}
\label{eq:app_backward_fpe}
    \partial_t \pi_t(x)
    =
    -\nabla\cdot\bigl(\pi_t(x)b_t^\leftarrow(x)\bigr)
    -
    \frac{\beta}{2}\Delta \pi_t(x).
\end{equation}

Since both Eq.~\eqref{eq:app_forward_fpe} and Eq.~\eqref{eq:app_backward_fpe} describe the same marginal density \(\pi_t\), their right-hand sides must be equal.
Equating them gives
\begin{equation}
    -\nabla\cdot\bigl(\pi_t b_t^\rightarrow\bigr)
    +
    \frac{\beta}{2}\Delta \pi_t
    =
    -\nabla\cdot\bigl(\pi_t b_t^\leftarrow\bigr)
    -
    \frac{\beta}{2}\Delta \pi_t .
\end{equation}
Rearranging terms yields
\begin{equation}
    \nabla\cdot\bigl(\pi_t b_t^\leftarrow\bigr)
    -
    \nabla\cdot\bigl(\pi_t b_t^\rightarrow\bigr)
    +
    \beta \Delta \pi_t
    =
    0 .
\end{equation}
Equivalently,
\begin{equation}
    \nabla\cdot
    \left(
        \pi_t b_t^\leftarrow
        -
        \pi_t b_t^\rightarrow
        +
        \beta \nabla \pi_t
    \right)
    =
    0 .
\end{equation}
Using \(\nabla \pi_t = \pi_t \nabla \log \pi_t\), we obtain
\begin{equation}
    \nabla\cdot
    \left[
        \pi_t
        \left(
            b_t^\leftarrow
            -
            b_t^\rightarrow
            +
            \beta \nabla \log \pi_t
        \right)
    \right]
    =
    0 .
\end{equation}
Therefore, the vector field inside the parentheses is determined up to a \(\pi_t\)-weighted divergence-free component.
That is, there exists a vector field \(h_t\) satisfying
\begin{equation}
    \nabla\cdot(\pi_t h_t)=0
\end{equation}
such that
\begin{equation}
    b_t^\leftarrow(x)
    -
    b_t^\rightarrow(x)
    +
    \beta \nabla \log \pi_t(x)
    =
    -h_t(x).
\end{equation}
Rearranging gives
\begin{equation}
\label{eq:app_bidirectional_consistency_general}
    b_t^\rightarrow(x) - b_t^\leftarrow(x)
    =
    \beta \nabla \log \pi_t(x) + h_t(x),
    \qquad
    \nabla\cdot(\pi_t h_t)=0 .
\end{equation}
This recovers Eq.~\eqref{eq:bidirectional_consistency_general} in the main text.

The additional field \(h_t\) is invisible at the level of marginal density evolution because it satisfies
\(\nabla\cdot(\pi_t h_t)=0\).
In other words, adding such a component changes the pointwise drift decomposition but does not change the induced evolution of \(\pi_t\).
Throughout this work, we focus on the canonical gradient representative by setting \(h_t=0\).
Under this choice, Eq.~\eqref{eq:app_bidirectional_consistency_general} reduces to
\begin{equation}
\label{eq:app_bidirectional_consistency}
    b_t^\rightarrow(x) - b_t^\leftarrow(x)
    =
    \beta \nabla \log \pi_t(x),
\end{equation}
which is Eq.~\eqref{eq:bidirectional_consistency} in the main text.

%%%%%%%%%%%%%%%%%%%%%%%%%%%%%%%%%%%%%%%%%%%%%%%%%%%%%%%%%%%%%%%%%%
\subsection{Full proof of Proposition~\ref{propos:d_component}}
\label{app:proof-d-component}

\begin{proof}
We prove that the antisymmetric component in the transport--osmotic decomposition is determined by the marginal score under the canonical representative.
By Definition~\ref{def:decomp}, we have
\begin{equation}
    d_t(x)
    =
    \frac{1}{2}
    \left(
        b_t^\rightarrow(x)
        -
        b_t^\leftarrow(x)
    \right).
\end{equation}
Under the canonical representative, the forward--backward drift relation from Eq.~\eqref{eq:bidirectional_consistency} gives
\begin{equation}
    b_t^\rightarrow(x)
    -
    b_t^\leftarrow(x)
    =
    \beta\nabla\log\pi_t(x).
\end{equation}
Substituting this identity into the definition of \(d_t\), we obtain
\begin{equation}
    d_t(x)
    =
    \frac{1}{2}
    \beta\nabla\log\pi_t(x)
    =
    \frac{\beta}{2}\nabla\log\pi_t(x).
\end{equation}
This proves Eq.~\eqref{eq:d_meaning}.
\end{proof}

%%%%%%%%%%%%%%%%%%%%%%%%%%%%%%%%%%%%%%%%%%%%%%%%%%%%%%%%%%%%%%%%%%
\subsection{Full proof of Proposition~\ref{propos:u_component}}
\label{app:proof-u-component}

\begin{proof}
We show that the symmetric component \(u_t\) governs the marginal density evolution through a deterministic continuity equation.
Recall the forward and backward marginal equations:
\begin{align}
    \partial_t \pi_t(x)
    &=
    -\nabla\cdot\bigl(\pi_t(x)b_t^\rightarrow(x)\bigr)
    +
    \frac{\beta}{2}\Delta \pi_t(x),
    \\
    \partial_t \pi_t(x)
    &=
    -\nabla\cdot\bigl(\pi_t(x)b_t^\leftarrow(x)\bigr)
    -
    \frac{\beta}{2}\Delta \pi_t(x).
\end{align}
Adding these two equations gives
\begin{equation}
    2\partial_t \pi_t(x)
    =
    -\nabla\cdot\bigl(\pi_t(x)b_t^\rightarrow(x)\bigr)
    -
    \nabla\cdot\bigl(\pi_t(x)b_t^\leftarrow(x)\bigr).
\end{equation}
Using the linearity of the divergence operator, we obtain
\begin{equation}
    2\partial_t \pi_t(x)
    =
    -\nabla\cdot
    \left[
        \pi_t(x)
        \bigl(
            b_t^\rightarrow(x)+b_t^\leftarrow(x)
        \bigr)
    \right].
\end{equation}
By Definition~\ref{def:decomp}, the symmetric component is
\begin{equation}
    u_t(x)
    =
    \frac{1}{2}
    \bigl(
        b_t^\rightarrow(x)+b_t^\leftarrow(x)
    \bigr).
\end{equation}
Therefore,
\begin{equation}
    2\partial_t \pi_t(x)
    =
    -2\nabla\cdot\bigl(\pi_t(x)u_t(x)\bigr).
\end{equation}
Dividing both sides by \(2\), we obtain
\begin{equation}
    \partial_t \pi_t(x)
    +
    \nabla\cdot\bigl(\pi_t(x)u_t(x)\bigr)
    =
    0,
\end{equation}
which proves Eq.~\eqref{eq:u_continuity_eq}.

We next derive the Lagrangian form of the same equation.
Expanding the divergence term in Eq.~\eqref{eq:u_continuity_eq} gives
\begin{equation}
    \partial_t \pi_t(x)
    +
    u_t(x)\cdot\nabla\pi_t(x)
    +
    \pi_t(x)\nabla\cdot u_t(x)
    =
    0.
\end{equation}
Assuming \(\pi_t(x)>0\), we divide by \(\pi_t(x)\):
\begin{equation}
    \frac{\partial_t\pi_t(x)}{\pi_t(x)}
    +
    u_t(x)\cdot
    \frac{\nabla\pi_t(x)}{\pi_t(x)}
    +
    \nabla\cdot u_t(x)
    =
    0.
\end{equation}
Using
\begin{equation}
    \partial_t\log\pi_t(x)
    =
    \frac{\partial_t\pi_t(x)}{\pi_t(x)},
    \qquad
    \nabla\log\pi_t(x)
    =
    \frac{\nabla\pi_t(x)}{\pi_t(x)},
\end{equation}
we obtain
\begin{equation}
    \partial_t\log\pi_t(x)
    +
    u_t(x)\cdot\nabla\log\pi_t(x)
    =
    -\nabla\cdot u_t(x).
\end{equation}
By the definition of the Lagrangian derivative along characteristics generated by \(u_t\),
\begin{equation}
    \frac{D}{Dt}f(x,t)
    :=
    \partial_t f(x,t)
    +
    u_t(x)\cdot\nabla f(x,t),
\end{equation}
we conclude that
\begin{equation}
    \frac{D}{Dt}\log\pi_t(x)
    =
    -\nabla\cdot u_t(x).
\end{equation}
This proves Eq.~\eqref{eq:u_meaning}.
\end{proof}

%%%%%%%%%%%%%%%%%%%%%%%%%%%%%%%%%%%%%%%%%%%%%%%%%%%%%%%%%%%%%%%%%%
%%%%%%%%%%%%%%%%%%%%%%%%%%%%%%%%%%%%%%%%%%%%%%%%%%%%%%%%%%%%%%%%%%
\section{Bridge Matching formulations}
\label{app:bm-formulations}

%%%%%%%%%%%%%%%%%%%%%%%%%%%%%%%%%%%%%%%%%%%%%%%%%%%%%%%%%%%%%%%%%%
\subsection{Relation to the Flow Matching objective}
\label{app:fm-bm-objectives}

Flow Matching trains a velocity field \(v_\theta\) to match a target velocity field \(v^*\) along a prescribed probability path.
Given samples \(x_t\sim \pi_t\) and time \(t\sim \mathcal{U}[0,1]\), the standard regression objective can be written as
\begin{equation}
\label{eq:app_fm_loss}
    \mathcal{L}_{\mathrm{FM}}
    =
    \mathbb{E}_{t,x_t}
    \left[
        \left\|
        v_\theta(x_t,t)-v^*(x_t,t)
        \right\|^2
    \right].
\end{equation}
Here, \(v^*(x_t,t)\) denotes the target vector field associated with the chosen probability path, such as a conditional linear interpolation path or a diffusion-style path.

Bridge Matching can be viewed as a decomposed counterpart of Flow Matching, where the single target velocity is replaced by transport and osmotic target fields.
Given target fields \(u^*(x,t)\) and \(d^*(x,t)\), we train two networks \(u_\theta\) and \(d_\phi\) using
\begin{equation}
\label{eq:app_bm_loss}
\mathcal{L}_{\mathrm{BM}}
=
\mathbb{E}_{t,x_t}
\left[
\|u_\theta(x_t,t)-u^*(x_t,t)\|^2
+
\lambda_d
\|d_\phi(x_t,t)-d^*(x_t,t)\|^2
\right].
\end{equation}
The coefficient \(\lambda_d>0\) controls the relative weight of the osmotic regression term.
When the learned components are recombined as
\begin{equation}
    v_{\theta,\phi}(x,t)
    =
    u_\theta(x,t)+d_\phi(x,t),
\end{equation}
the resulting field can be used for forward sample generation.
More generally, evaluating
\begin{equation}
    v_{\theta,\phi}^{(\lambda_u,\lambda_d)}(x,t)
    =
    \lambda_u u_\theta(x,t)
    +
    \lambda_d d_\phi(x,t)
\end{equation}
allows the relative contribution of transport and osmotic effects to be adjusted at sampling time.

%%%%%%%%%%%%%%%%%%%%%%%%%%%%%%%%%%%%%%%%%%%%%%%%%%%%%%%%%%%%%%%%%%
\subsection{Marginal Bridge Matching}
\label{app:marginal-bridge-matching}

We present the details of marginal formulation of Bridge Matching and its implementation here, as additional information to Section \ref{sec:implementation}.
This formulation directly follows the transport--osmotic decomposition at the level of the intermediate marginal distribution.
Given a probability path \(\{\pi_t\}_{t\in[0,1]}\), Marginal Bridge Matching aims to learn the marginal transport field \(u_t\) and the osmotic field \(d_t\) associated with this path.

The transport component is defined as the deterministic velocity field that realizes the marginal evolution through the continuity equation:
\begin{equation}
\label{eq:app_mbm_transport_continuity}
    \partial_t \pi_t(x)
    +
    \nabla\cdot\bigl(\pi_t(x)u_t^*(x)\bigr)
    =
    0.
\end{equation}
Thus, \(u_t^*\) represents a deterministic field that carries the marginal density \(\pi_t\) along the prescribed probability path.

The osmotic component is defined from the marginal score.
Under the canonical representative, Proposition~\ref{propos:d_component} gives
\begin{equation}
\label{eq:app_mbm_osmotic_target}
    d_t^*(x)
    =
    \frac{\beta_t}{2}
    \nabla_x \log \pi_t(x).
\end{equation}
Therefore, if the marginal score \(\nabla_x\log\pi_t(x)\) is available, the transport and osmotic targets can be directly constructed from the marginal path.

In practice, however, the marginal density \(\pi_t\) is rarely available in closed form.
We therefore approximate its score from samples
\(\{x_t^{(i)}\}_{i=1}^B\) drawn from the intermediate marginal distribution.
A simple choice is a kernel density estimator (KDE).
For a Gaussian kernel with bandwidth \(h>0\), define the pairwise squared distances
\begin{equation}
    D_{ij}
    =
    \left\|x_t^{(i)}-x_t^{(j)}\right\|^2 .
\end{equation}
The normalized leave-one-out kernel weights are
\begin{equation}
\label{eq:app_kde_weights}
w_{ij}
=
\frac{
\exp\!\left(-\frac{D_{ij}}{2h^2}\right)
}{
\sum_{k\neq i}
\exp\!\left(-\frac{D_{ik}}{2h^2}\right)
},
\qquad j\neq i.
\end{equation}
The marginal score at \(x_t^{(i)}\) can then be approximated by
\begin{equation}
\label{eq:app_kde_score}
\nabla_x \log \hat{\pi}_t(x_t^{(i)})
\approx
\frac{
\sum_{j\neq i} w_{ij}x_t^{(j)} - x_t^{(i)}
}{h^2}.
\end{equation}
This gives the empirical osmotic target
\begin{equation}
\label{eq:app_mbm_kde_d}
    d_t^*(x_t^{(i)})
    =
    \frac{\beta_t}{2}
    \nabla_x \log \hat{\pi}_t(x_t^{(i)}).
\end{equation}

The resulting Marginal Bridge Matching objective is
\begin{equation}
\label{eq:app_mbm_loss}
\mathcal{L}_{\mathrm{MBM}}
=
\mathbb{E}_{t,x_t\sim\pi_t}
\left[
\left\|u_\theta(x_t,t)-u_t^*(x_t)\right\|^2
+
\lambda_d
\left\|d_\phi(x_t,t)-d_t^*(x_t)\right\|^2
\right].
\end{equation}

This construction directly follows the marginal definition of the decomposition:
the transport component is chosen to satisfy the continuity equation, while the osmotic component is determined by the marginal score.
It is therefore applicable to general probability paths as long as intermediate samples and transport targets can be obtained.
However, it requires estimating the marginal score from samples, which can introduce approximation error, especially in high-dimensional spaces.

%%%%%%%%%%%%%%%%%%%%%%%%%%%%%%%%%%%%%%%%%%%%%%%%%%%%%%%%%%%%%%%%%%
\subsection{Conditional Bridge Matching}
\label{app:conditional-bridge-matching}

We next present a conditional formulation that provides a tractable alternative to the marginal construction.
Instead of directly estimating the marginal score \(\nabla_x\log\pi_t(x)\), Conditional Bridge Matching specifies the intermediate path through a conditional distribution with tractable density.

Let \(z\sim p(z)\) be an auxiliary conditioning variable, such as endpoint samples, latent variables, noise variables, or any variable used to define the interpolation path.
For each \(z\), we define a conditional stochastic interpolation
\begin{equation}
    x_t \sim p_t(x\mid z),
    \qquad t\in[0,1],
\end{equation}
where the conditional density \(p_t(x\mid z)\) and its score are assumed to be available in closed form.
The marginal path is obtained by averaging over the conditioning variable:
\begin{equation}
    \pi_t(x)
    =
    \int p_t(x\mid z)p(z)\,dz.
\end{equation}

Given the conditional path \(p_t(x\mid z)\), we define conditional target fields \(u_t^*(x_t,z)\) and \(d_t^*(x_t,z)\).
Let \(v_t^*(x_t,z)\) denote the total conditional velocity associated with the prescribed conditional evolution.
For example, if the conditional interpolation is generated by a differentiable map
\begin{equation}
    x_t = I_t(z,\varepsilon),
\end{equation}
where \(\varepsilon\) is an auxiliary noise variable, then a natural total conditional velocity is
\begin{equation}
\label{eq:app_cbm_velocity_target}
    v_t^*(x_t,z)
    =
    \partial_t I_t(z,\varepsilon).
\end{equation}
The osmotic target is defined from the conditional score:
\begin{equation}
\label{eq:app_cbm_conditional_score}
    d_t^*(x_t,z)
    =
    \frac{\beta_t}{2}
    \nabla_{x_t}\log p_t(x_t\mid z).
\end{equation}
The transport target is then defined by subtracting the osmotic target from the total conditional velocity:
\begin{equation}
\label{eq:app_cbm_transport_target}
    u_t^*(x_t,z)
    =
    v_t^*(x_t,z)-d_t^*(x_t,z).
\end{equation}
This construction ensures that
\begin{equation}
    v_t^*(x_t,z)
    =
    u_t^*(x_t,z)+d_t^*(x_t,z).
\end{equation}

The resulting Conditional Bridge Matching objective is
\begin{equation}
\label{eq:app_cbm_loss}
\mathcal{L}_{\mathrm{CBM}}
=
\mathbb{E}_{t,z,x_t\sim p_t(\cdot\mid z)}
\left[
\left\|u_\theta(x_t,t)-u_t^*(x_t,z)\right\|^2
+
\lambda_d
\left\|d_\phi(x_t,t)-d_t^*(x_t,z)\right\|^2
\right].
\end{equation}

This conditional formulation avoids explicit estimation of the marginal score by using a tractable conditional score.
As a result, it gives a simple and efficient conditional approximation to the transport--osmotic decomposition.
However, because the targets are defined conditionally, the learned fields approximate the marginal decomposition only after averaging over the chosen conditional construction.
The quality of this approximation therefore depends on the choice of conditional path.

%%%%%%%%%%%%%%%%%%%%%%%%%%%%%%%%%%%%%%%%%%%%%%%%%%%%%%%%%%%%%%%%%%
\subsection{Diffusion-style conditional bridge}
\label{app:diffusion-style-cbm}

A common conditional construction uses a Gaussian stochastic interpolation.
Given \(x_1\sim\pi_1\) and \(\varepsilon\sim\mathcal{N}(0,I)\), define
\begin{equation}
\label{eq:app_diffusion_conditional_path}
    x_t
    =
    \alpha(t)x_1
    +
    \sigma(t)\varepsilon,
\end{equation}
where \(\alpha(t)\) and \(\sigma(t)\) are smooth scalar schedules.
This induces the conditional density
\begin{equation}
\label{eq:app_diffusion_conditional_density}
    p_t(x_t\mid x_1)
    =
    \mathcal{N}
    \left(
        \alpha(t)x_1,
        \sigma^2(t)I
    \right).
\end{equation}

The conditional score is available in closed form:
\begin{equation}
\label{eq:app_diffusion_conditional_score}
    \nabla_{x_t}\log p_t(x_t\mid x_1)
    =
    -
    \frac{x_t-\alpha(t)x_1}{\sigma^2(t)}.
\end{equation}
Therefore, the conditional osmotic target is
\begin{equation}
\label{eq:app_diffusion_conditional_d}
    d_t^*(x_t,x_1)
    =
    -
    \frac{\beta_t}{2}
    \frac{x_t-\alpha(t)x_1}{\sigma^2(t)}.
\end{equation}

The total conditional velocity is obtained by differentiating the interpolation:
\begin{equation}
\label{eq:app_diffusion_conditional_v}
    v_t^*(x_t,x_1)
    =
    \partial_t
    \left(
        \alpha(t)x_1+\sigma(t)\varepsilon
    \right)
    =
    \dot{\alpha}(t)x_1+\dot{\sigma}(t)\varepsilon.
\end{equation}
Equivalently, since
\[
\varepsilon=\frac{x_t-\alpha(t)x_1}{\sigma(t)},
\]
this target can be written as
\begin{equation}
\label{eq:app_diffusion_conditional_v_xt}
    v_t^*(x_t,x_1)
    =
    \dot{\alpha}(t)x_1
    +
    \frac{\dot{\sigma}(t)}{\sigma(t)}
    \left(
        x_t-\alpha(t)x_1
    \right).
\end{equation}
The corresponding conditional transport target is then
\begin{equation}
\label{eq:app_diffusion_conditional_u}
    u_t^*(x_t,x_1)
    =
    v_t^*(x_t,x_1)-d_t^*(x_t,x_1).
\end{equation}
Thus, the diffusion-style conditional bridge satisfies
\begin{equation}
    v_t^*(x_t,x_1)
    =
    u_t^*(x_t,x_1)+d_t^*(x_t,x_1).
\end{equation}

In our experiments, one possible choice is the trigonometric schedule
\begin{equation}
    \alpha(t)=\sin\left(\frac{\pi}{2}t\right),
    \qquad
    \sigma(t)=\cos\left(\frac{\pi}{2}t\right),
\end{equation}
which smoothly connects a noise-dominated distribution at \(t=0\) to the data endpoint at \(t=1\).
This construction provides closed-form conditional transport and osmotic targets, avoiding explicit marginal score estimation.
%%%%%%%%%%%%%%%%%%%%%%%%%%%%%%%%%%%%%%%%%%%%%%%%%%%%%%%%%%%%
%%%%%%%%%%%%%%%%%%%%%%%%%%%%%%%%%%%%%%%%%%%%%%%%%%%%%%%%%%%%
\section{Implementation details for FM and BM targets}
\label{app:target-details}

We provide implementation details for the FM and BM target constructions used in the 2D generation experiments.
We evaluate Bridge Matching on 2D transport tasks involving several source--target pairs, including checkerboard, moons, and Gaussian mixture distributions.
For each task, we sample source points \(x_0\sim\pi_0\), target points \(x_1\sim\pi_1\), and time \(t\sim\mathcal{U}[0,1]\).
We compare six target constructions:
CFM-Linear, CFM-Diffusion, CBM-Linear, CBM-Diffusion, MBM-Linear, and MBM-Diffusion.

All methods are implemented within the same target interface.
Each target construction returns an intermediate sample \(x_t\), a transport target \(u_t^*\), and an osmotic target \(d_t^*\).
For pure CFM baselines, we set
\begin{equation}
    d_t^*=0,
    \qquad
    u_t^*=v_t^*,
\end{equation}
where \(v_t^*\) is the standard Flow Matching target velocity.
For BM variants, we construct \(d_t^*\) from either a conditional score or a marginal score estimate, and define
\begin{equation}
    u_t^*(x_t)
    =
    v_t^*(x_t)-d_t^*(x_t),
\end{equation}
so that the recombined field satisfies
\begin{equation}
    v_t^*(x_t)
    =
    u_t^*(x_t)+d_t^*(x_t).
\end{equation}
In implementation, the scalar coefficient used for the osmotic target is denoted by \(\beta_{\mathrm{impl}}\).
This coefficient absorbs constant factors such as the theoretical \(1/2\) in the relation
\(d_t=(\beta_t/2)\nabla\log\pi_t\), and is treated as a tunable scale parameter.

\paragraph{CFM-Linear.}
For the linear CFM baseline, we use the standard conditional affine interpolation
\begin{equation}
    x_t
    =
    t x_1 + \sigma(t)x_0,
\end{equation}
where
\begin{equation}
    \sigma(t)
    =
    1-(1-\sigma_{\min})t.
\end{equation}
The corresponding CFM target velocity is
\begin{equation}
    v_t^*
    =
    x_1
    +
    \frac{\dot{\sigma}(t)}{\sigma(t)}
    \bigl(x_t-tx_1\bigr).
\end{equation}
For this baseline, we set
\begin{equation}
    u_t^*=v_t^*,
    \qquad
    d_t^*=0.
\end{equation}

\paragraph{CFM-Diffusion.}
For the diffusion CFM baseline, we use the variance-preserving affine probability path
\begin{equation}
    x_t
    =
    \alpha(t)x_1+\sigma(t)x_0,
\end{equation}
where \(\alpha(t)\) and \(\sigma(t)\) are determined by the VP scheduler.
The target velocity \(v_t^*\) is given by the time derivative of the sampled path, as in the standard affine Flow Matching construction.
Again, for the pure CFM baseline, we set
\begin{equation}
    u_t^*=v_t^*,
    \qquad
    d_t^*=0.
\end{equation}

\paragraph{CBM-Linear.}
For the linear conditional BM target, we construct a stochastic tube around the paired linear path.
The mean path is
\begin{equation}
    m_t
    =
    (1-t)x_0+t x_1.
\end{equation}
We then sample
\begin{equation}
    x_t
    =
    m_t+\sigma(t)\varepsilon,
    \qquad
    \varepsilon\sim\mathcal{N}(0,I),
\end{equation}
where
\begin{equation}
    \sigma(t)
    =
    \sqrt{t(1-t)}
\end{equation}
with a numerical floor \(\sigma_{\min}\) for stability.
This produces a tube-like stochastic path around the linear interpolation, while still preserving the same linear transport geometry.

The conditional score of the Gaussian tube is
\begin{equation}
    \nabla_{x_t}\log p_t(x_t\mid x_0,x_1)
    =
    -
    \frac{x_t-m_t}{\sigma^2(t)}.
\end{equation}
The osmotic target is implemented as
\begin{equation}
    d_t^*
    =
    -\beta_{\mathrm{impl}}
    \frac{x_t-m_t}{\sigma^2(t)}.
\end{equation}
The full conditional path velocity is
\begin{equation}
    v_t^*
    =
    x_1-x_0
    +
    \frac{\dot{\sigma}(t)}{\sigma(t)}
    \bigl(x_t-m_t\bigr),
\end{equation}
and the transport target is defined by
\begin{equation}
    u_t^*
    =
    v_t^*-d_t^*.
\end{equation}

\paragraph{CBM-Diffusion.}
For the diffusion conditional BM target, we use the same VP affine probability path as CFM-Diffusion:
\begin{equation}
    x_t
    =
    \alpha(t)x_1+\sigma(t)x_0.
\end{equation}
The total target velocity \(v_t^*\) is exactly the affine Flow Matching target for this path.
The conditional score is
\begin{equation}
    \nabla_{x_t}\log p_t(x_t\mid x_1)
    =
    -
    \frac{x_t-\alpha(t)x_1}{\sigma^2(t)}.
\end{equation}
We define
\begin{equation}
    d_t^*
    =
    \beta_{\mathrm{impl}}
    \nabla_{x_t}\log p_t(x_t\mid x_1),
\end{equation}
and set
\begin{equation}
    u_t^*
    =
    v_t^*-d_t^*.
\end{equation}
Thus, CBM-Diffusion uses the same probability path and total velocity target as CFM-Diffusion, but decomposes the target into transport and osmotic components.

\paragraph{MBM-Linear.}
For the linear marginal BM target, we use the deterministic linear interpolation
\begin{equation}
    x_t
    =
    (1-t)x_0+t x_1.
\end{equation}
The corresponding path velocity is
\begin{equation}
    v_t^*
    =
    x_1-x_0.
\end{equation}
Unlike CBM-Linear, which uses a conditional Gaussian tube score, MBM-Linear estimates the marginal score \(\nabla\log\pi_t(x_t)\) from the batch using KDE.
Given a mini-batch \(\{x_t^{(i)}\}_{i=1}^B\), we compute pairwise distances
\begin{equation}
    D_{ij}
    =
    \left\|x_t^{(i)}-x_t^{(j)}\right\|^2,
\end{equation}
and define leave-one-out Gaussian kernel weights
\begin{equation}
    w_{ij}
    =
    \frac{
    \exp\!\left(-D_{ij}/(2h^2)\right)
    }{
    \sum_{k\neq i}\exp\!\left(-D_{ik}/(2h^2)\right)
    },
    \qquad j\neq i.
\end{equation}
The marginal score is approximated by
\begin{equation}
    \nabla\log\hat{\pi}_t(x_t^{(i)})
    \approx
    \frac{
    \sum_{j\neq i}w_{ij}x_t^{(j)}-x_t^{(i)}
    }{h^2}.
\end{equation}
We then set
\begin{equation}
    d_t^*(x_t^{(i)})
    =
    \beta_{\mathrm{impl}}
    \nabla\log\hat{\pi}_t(x_t^{(i)}),
    \qquad
    u_t^*(x_t^{(i)})
    =
    v_t^*(x_t^{(i)})-d_t^*(x_t^{(i)}).
\end{equation}

\paragraph{MBM-Diffusion.}
For the diffusion marginal BM target, we use the same VP affine probability path as CFM-Diffusion and CBM-Diffusion:
\begin{equation}
    x_t
    =
    \alpha(t)x_1+\sigma(t)x_0.
\end{equation}
The total target velocity \(v_t^*\) is again the standard affine Flow Matching target.
However, instead of using the conditional Gaussian score around each individual \(x_1\), we estimate the marginal score from the batch using the same KDE approximation as in MBM-Linear:
\begin{equation}
    d_t^*(x_t^{(i)})
    =
    \beta_{\mathrm{impl}}
    \nabla\log\hat{\pi}_t(x_t^{(i)}).
\end{equation}
The transport target is then
\begin{equation}
    u_t^*(x_t^{(i)})
    =
    v_t^*(x_t^{(i)})-d_t^*(x_t^{(i)}).
\end{equation}
This variant retains the diffusion-style probability path while using a marginal, rather than conditional, approximation of the osmotic component.

\paragraph{Sampling with recombined dynamics.}
After training, samples are generated by recombining the learned fields:
\begin{equation}
    v_{\theta,\phi}^{(\lambda_u,\lambda_d)}(x,t)
    =
    \lambda_u u_\theta(x,t)
    +
    \lambda_d d_\phi(x,t).
\end{equation}
Unless otherwise stated, we fix \(\lambda_u=1\) and vary \(\lambda_d\) to evaluate the effect of the learned osmotic component.
This allows us to test whether the osmotic field improves endpoint matching and to analyze how stochastic density-shaping effects influence generation.
%%%%%%%%%%%%%%%%%%%%%%%%%%%%%%%%%%%%%%%%%%%%%%%%%%%%%%%%%%%%
%%%%%%%%%%%%%%%%%%%%%%%%%%%%%%%%%%%%%%%%%%%%%%%%%%%%%%%%%%%%
\section{Computational resources}
\label{app:computational-resources}

We report the computational environments used to run the experiments for reproducibility. 
Experiments were conducted on two internal workstations. 
The first workstation was equipped with a single NVIDIA GeForce RTX 5090 GPU with 32GB memory, an Intel Core Ultra 9 285 CPU with 24 logical CPUs, and 62GB system memory. 
The software environment on this machine used NVIDIA driver 570.211.01, CUDA toolkit 12.8.93, Python 3.9.23, PyTorch 2.8.0+cu128, and cuDNN 91002.

The second workstation was equipped with two NVIDIA RTX 6000 Ada Generation GPUs, each with 48GB memory, an Intel Xeon w9-3495X CPU with 56 physical cores and 112 logical CPUs, and 503GB system memory. 
This machine used NVIDIA driver 555.58.02. 
Although the CUDA toolkit compiler \texttt{nvcc} was not available in the active environment on this machine, PyTorch reported CUDA 12.8 support through PyTorch 2.8.0+cu128, with cuDNN 91002.

Both machines used shared network-mounted storage. 
The home directory was mounted on a 4.0TB shared filesystem, and the main shared data directory was mounted on a 29TB filesystem. 
The \texttt{boeuf} workstation additionally provided a local 894GB data disk. 
The 2D experiments were lightweight and were run on the same computational environment, whereas the image-generation experiments were GPU-intensive. 
For representative image-generation runs on \texttt{souris}, GPU memory usage was approximately 26GB out of 32GB.
Unless otherwise stated, the reported experiments used the Python and PyTorch environment described above.

%%%%%%%%%%%%%%%%%%%%%%%%%%%%%%%%%%%%%%%%%%%%%%%%%%%%%%%%%%%%
%%%%%%%%%%%%%%%%%%%%%%%%%%%%%%%%%%%%%%%%%%%%%%%%%%%%%%%%%%%%
\section{Implementation details for 2D generation}
\label{app:2d-generation-details}

%%%%%%%%%%%%%%%%%%%%%%%%%%%%%%%%%%%%%%%%%%%%%%%%%%%%%%%%%%%%
\subsection{2D synthetic datasets}
\label{app:2d-datasets}

We evaluate 2D generation on several synthetic distributions.
All datasets are defined in \(\mathbb{R}^2\), and samples are drawn independently during training and evaluation.
For visualization, we use a square plotting range computed from samples so that the two axes share the same scale.

\paragraph{Gaussian.}
The Gaussian dataset is a diagonal Gaussian distribution:
\begin{equation}
    x\sim\mathcal{N}(\mu,\operatorname{diag}(\sigma^2)).
\end{equation}
By default, we use
\begin{equation}
    \mu=(0,0),
    \qquad
    \sigma=(1,1).
\end{equation}

\paragraph{Moons.}
The moons dataset consists of two interleaving half circles generated using the standard two-moons distribution.
Unless otherwise stated, we use Gaussian perturbation noise with standard deviation \(0.05\).

\paragraph{Gaussian mixture.}
The Gaussian mixture dataset consists of four isotropic Gaussian components with standard deviation \(0.5\).
The component means are
\begin{equation}
    (0,-2),\qquad
    (0,0),\qquad
    (2,2),\qquad
    (-2,2).
\end{equation}
Each component is sampled with equal probability.

\paragraph{Checkerboard.}
The checkerboard dataset follows the common 2D checkerboard construction used in flow-based generative modeling.
We first sample
\begin{equation}
    z_1 \sim \mathcal{U}[-2,2],
\end{equation}
and construct the second coordinate by alternating the vertical offset according to the integer part of \(z_1\):
\begin{equation}
    z_2
    =
    u - 2k + \lfloor z_1 \rfloor \bmod 2,
    \qquad
    u\sim\mathcal{U}[0,1],\quad
    k\sim\mathrm{Uniform}\{0,1\}.
\end{equation}
The final sample is scaled by \(1/0.45\):
\begin{equation}
    x
    =
    \frac{1}{0.45}(z_1,z_2).
\end{equation}

%%%%%%%%%%%%%%%%%%%%%%%%%%%%%%%%%%%%%%%%%%%%%%%%%%%%%%%%%%%%%%%%%%%%%%%%
\subsection{Training configuration}
\label{app:2d-training-config}

For all 2D experiments, we train two separate neural networks \(u_\theta\) and \(d_\phi\) to model the transport and osmotic components, respectively.
Both networks use the same multilayer perceptron architecture.
Given an intermediate sample \(x_t\in\mathbb{R}^2\) and time \(t\in[0,1]\), the input is the concatenation \([x_t,t]\).
The network consists of three hidden layers with SiLU activations and a final linear layer mapping back to \(\mathbb{R}^2\):
\begin{equation}
    [x_t,t]
    \rightarrow
    \mathrm{Linear}
    \rightarrow
    \mathrm{SiLU}
    \rightarrow
    \mathrm{Linear}
    \rightarrow
    \mathrm{SiLU}
    \rightarrow
    \mathrm{Linear}
    \rightarrow
    \mathrm{SiLU}
    \rightarrow
    \mathrm{Linear}.
\end{equation}
Unless otherwise stated, the hidden dimension is set to \(512\).

At each training iteration, we sample a mini-batch of source points \(x_0\sim\pi_0\), target points \(x_1\sim\pi_1\), and times \(t\sim\mathcal{U}[0,1]\).
The selected target construction returns the intermediate point \(x_t\), transport target \(u_t^*\), and osmotic target \(d_t^*\).
The two networks are optimized with the mini-batch regression objective
\begin{equation}
    \mathcal{L}
    =
    \mathbb{E}_{t,x_t}
    \left[
    \left\|
    u_\theta(x_t,t)-u_t^*
    \right\|^2
    +
    \lambda_d
    \left\|
    d_\phi(x_t,t)-d_t^*
    \right\|^2
    \right].
\end{equation}
We use AdamW with learning rate \(10^{-3}\), batch size \(4096\), and train for \(100{,}000\) iterations.
The default random seed is \(42\).

The main hyperparameters used in the 2D experiments are summarized in Table \ref{tab:app_2d_training_config}.
\begin{table}[h]
\centering
\caption{Default training configuration for 2D experiments.}
\label{tab:app_2d_training_config}
\begin{tabular}{ll}
\toprule
Hyperparameter & Value \\
\midrule
Optimizer & AdamW \\
Learning rate & \(10^{-3}\) \\
Batch size & \(4096\) \\
Training iterations & \(100{,}000\) \\
Hidden dimension & \(512\) \\
Activation & SiLU \\
Loss weight \(\lambda_d\) & \(1.0\) \\
Default seed & \(42\) \\
Logging interval & \(2000\) iterations \\
\bottomrule
\end{tabular}
\end{table}

For diffusion-based targets, we use a default osmotic scale \(\beta_{\mathrm{impl}}=0.01\) and numerical floor \(\sigma_{\min}=0.05\).
For linear tube-based targets, we use a larger osmotic scale and floor when needed for numerical stability; in our implementation, typical values are \(\beta_{\mathrm{impl}}=0.1\) and \(\sigma_{\min}=0.1\).
The small time threshold \(t_{\epsilon}=10^{-2}\) is used where applicable to avoid numerical instability near the endpoints of the interpolation path.

After training, we evaluate the relative magnitude of the learned components by computing
\begin{equation}
    \mathbb{E}\left[\|u_\theta(x_t,t)\|\right],
    \qquad
    \mathbb{E}\left[\|d_\phi(x_t,t)\|\right],
\end{equation}
on a held-out mini-batch of intermediate samples.
We also report the ratio
\begin{equation}
    \frac{
    \mathbb{E}\left[\|d_\phi(x_t,t)\|\right]
    }{
    \mathbb{E}\left[\|u_\theta(x_t,t)\|\right]+\epsilon
    },
\end{equation}
where \(\epsilon=10^{-8}\) is used for numerical stability.
This diagnostic helps quantify the relative contribution of the learned osmotic component.

%%%%%%%%%%%%%%%%%%%%%%%%%%%%%%%%%%%%%%%%%%%%%%%%%%%%%%%%%%%%%%%%%%%%%%%%
\subsection{Evaluation configuration}
\label{app:2d-eval-config}

For 2D evaluation, we load the trained networks \(u_\theta\) and \(d_\phi\) from the saved checkpoint and generate samples by integrating recombined dynamics.
For forward generation from the source distribution \(\pi_0\) to the target distribution \(\pi_1\), we use
\begin{equation}
    v_{\theta,\phi}^{\mathrm{fwd}}(x,t)
    =
    \lambda_u u_\theta(x,t)
    +
    \lambda_d d_\phi(x,t).
\end{equation}
Unless otherwise stated, we fix \(\lambda_u=1\) and vary \(\lambda_d\) to study the effect of the osmotic component.

For backward generation, we use the reverse-oriented recombination
\begin{equation}
    v_{\theta,\phi}^{\mathrm{bwd}}(x,t)
    =
    -
    \left[
    \lambda_u u_\theta(x,1-t)
    -
    \lambda_d d_\phi(x,1-t)
    \right].
\end{equation}
This corresponds to integrating the backward dynamics from the target distribution to the source distribution, using the decomposition convention
\begin{equation}
    b_t^\rightarrow = u_t+d_t,
    \qquad
    b_t^\leftarrow = u_t-d_t.
\end{equation}

We solve the resulting ODE using the midpoint method with step size \(0.01\).
For sampling visualizations, we generate \(5\times 10^6\) samples and record \(50\) intermediate states along the trajectory.
For quantitative metrics, we use \(10{,}000\) generated samples and compare them with \(10{,}000\) target samples.
The default evaluation configuration is summarized in Table \ref{tab:app_2d_eval_config}.
\begin{table}[h]
\centering
\caption{Default evaluation configuration for 2D experiments.}
\label{tab:app_2d_eval_config}
\begin{tabular}{ll}
\toprule
Hyperparameter & Value \\
\midrule
Sampling method & Midpoint \\
Step size & \(0.01\) \\
Absolute tolerance & \(10^{-5}\) \\
Relative tolerance & \(10^{-5}\) \\
Visualization samples & \(5\times 10^6\) \\
Visualization steps & \(50\) \\
Metric samples & \(10{,}000\) \\
Default \(\lambda_u\) & \(1.0\) \\
Default \(\lambda_d\) & \(1.0\) \\
Default seed & \(42\) \\
\bottomrule
\end{tabular}
\end{table}

\paragraph{Quantitative metrics.}
For 2D experiments, we report two distributional metrics: a 2D Fréchet distance and an RBF-kernel maximum mean discrepancy.

The 2D Fréchet distance is computed directly on the sample coordinates.
Given real samples with empirical mean and covariance \((\mu_r,\Sigma_r)\) and generated samples with \((\mu_g,\Sigma_g)\), we compute
\begin{equation}
    \mathrm{FID}_{2D}
    =
    \|\mu_r-\mu_g\|_2^2
    +
    \operatorname{Tr}
    \left(
    \Sigma_r+\Sigma_g
    -
    2(\Sigma_r^{1/2}\Sigma_g\Sigma_r^{1/2})^{1/2}
    \right).
\end{equation}
Although this metric is analogous to FID, it is computed in the original 2D coordinate space rather than in an image-feature space.

We also compute the squared maximum mean discrepancy with an RBF kernel:
\begin{equation}
    \mathrm{MMD}^2
    =
    \frac{1}{n(n-1)}
    \sum_{i\neq j}k(x_i,x_j)
    +
    \frac{1}{m(m-1)}
    \sum_{i\neq j}k(y_i,y_j)
    -
    \frac{2}{nm}
    \sum_{i,j}k(x_i,y_j),
\end{equation}
where \(\{x_i\}_{i=1}^n\) are real samples, \(\{y_j\}_{j=1}^m\) are generated samples, and
\begin{equation}
    k(x,y)
    =
    \exp
    \left(
    -\frac{\|x-y\|^2}{2\sigma^2}
    \right).
\end{equation}
The bandwidth \(\sigma^2\) is selected using the median heuristic on pairwise real--generated distances.

\paragraph{Uncertainty reporting for 2D metrics.}
For the 2D experiments, \(\mathrm{FID}_{2D}\) and \(\mathrm{MMD}^2\) are computed from finite generated and target sample sets using the same evaluation protocol across methods. 
We report these metrics as single-run distributional estimates and do not claim statistical significance across independently trained seeds. 
When multiple stochastic samples are involved, variability comes from the generated and target sample draws under the fixed model and sampling configuration. 
Trajectory plots and vector-field visualizations are used as qualitative diagnostics rather than statistical-significance evidence.

\paragraph{Field decomposition visualization.}
To visualize the learned decomposition, we evaluate \(u_\theta\) and \(d_\phi\) on a uniform 2D grid over the region covering source and target samples.
For each selected time value, we plot the magnitude of each field as a heatmap and visualize the field direction using streamlines.
Unless otherwise stated, we use a \(45\times45\) grid and visualize \(5\) time steps.
The plotting range is estimated from \(20{,}000\) source samples and \(20{,}000\) target samples.
%%%%%%%%%%%%%%%%%%%%%%%%%%%%%%%%%%%%%%%%%%%%%%%%%%%%%%%%%%%%
%%%%%%%%%%%%%%%%%%%%%%%%%%%%%%%%%%%%%%%%%%%%%%%%%%%%%%%%%%%%
\section{Implementation details for image generation}
\label{app:image-generation-details}

We evaluate image generation on CIFAR-10, ImageNet-32, and ImageNet-64.
These datasets provide progressively more challenging image-generation settings, ranging from small-scale natural images to downsampled ImageNet at higher spatial resolutions.
All images are represented as RGB images and normalized to the range used by the training pipeline.

\paragraph{Configuration reporting.}
In the following sections, we report the experiment-designated configuration values used for each image-generation setting.
These tables list the arguments that were explicitly set or that are most relevant for reproducing the reported comparisons.
Unless otherwise stated, all unspecified arguments follow the default values defined in the training and evaluation argument parsers.
For reproducibility, each run also stores the full parsed argument namespace as an \texttt{args.json} file in the corresponding output directory, including both user-specified arguments and parser defaults.
Therefore, the appendix tables should be interpreted as a compact summary of the main experimental settings rather than an exhaustive dump of every command-line argument.

For each dataset, CFM-Diffusion and CBM-Diffusion use the same data pipeline, architecture, optimizer, EMA setting, timestep sampling strategy, solver, and evaluation protocol unless a method-specific argument is explicitly listed.
The main methodological difference is the target construction:
CFM-Diffusion directly regresses the diffusion-path Flow Matching velocity, whereas CBM-Diffusion learns separate transport and osmotic components and recombines them at sampling time.

%%%%%%%%%%%%%%%%%%%%%%%%%%%%%%%%%%%%%%%%%%%%%%%%%%%%%%%%%%%%
\subsection{Image-generation Metrics}
\label{app:image-generation-metrics}

We evaluate generated images using Fréchet Inception Distance (FID), Kernel Inception Distance (KID), and Inception Score (IS).
All three metrics are computed using features or predictions from a pretrained Inception network.

\paragraph{Fréchet Inception Distance.}
FID compares the generated and real image distributions in Inception feature space.
Let \(f(x)\) denote the Inception feature representation of image \(x\).
Given real images and generated images, we compute their empirical feature means and covariances:
\begin{equation}
    \mu_r = \mathbb{E}_{x\sim p_{\mathrm{data}}}[f(x)],
    \qquad
    \Sigma_r = \operatorname{Cov}_{x\sim p_{\mathrm{data}}}[f(x)],
\end{equation}
\begin{equation}
    \mu_g = \mathbb{E}_{x\sim p_{\theta}}[f(x)],
    \qquad
    \Sigma_g = \operatorname{Cov}_{x\sim p_{\theta}}[f(x)].
\end{equation}
The FID is computed as
\begin{equation}
    \mathrm{FID}
    =
    \|\mu_r-\mu_g\|_2^2
    +
    \operatorname{Tr}
    \left(
        \Sigma_r+\Sigma_g
        -
        2(\Sigma_r^{1/2}\Sigma_g\Sigma_r^{1/2})^{1/2}
    \right).
\end{equation}
Lower FID indicates that the generated image distribution is closer to the real image distribution in Inception feature space.

\paragraph{Kernel Inception Distance.}
KID measures the discrepancy between real and generated image distributions using maximum mean discrepancy (MMD) in Inception feature space.
Let \(\{f(x_i)\}_{i=1}^m\) be real image features and \(\{f(y_j)\}_{j=1}^n\) be generated image features.
KID is computed as an unbiased estimate of squared MMD:
\begin{equation}
    \mathrm{KID}
    =
    \frac{1}{m(m-1)}
    \sum_{i\neq j}
    k(f(x_i),f(x_j))
    +
    \frac{1}{n(n-1)}
    \sum_{i\neq j}
    k(f(y_i),f(y_j))
    -
    \frac{2}{mn}
    \sum_{i,j}
    k(f(x_i),f(y_j)).
\end{equation}
Following the standard KID definition, we use a polynomial kernel:
\begin{equation}
    k(a,b)
    =
    \left(
        \frac{1}{d}a^\top b + 1
    \right)^3,
\end{equation}
where \(d\) is the feature dimension.
Lower KID indicates better agreement between real and generated image distributions.

\paragraph{Inception Score.}
Inception Score evaluates both sample quality and diversity using the class-prediction distribution of a pretrained Inception classifier.
Let \(p(y\mid x)\) be the predicted label distribution for generated image \(x\), and let
\begin{equation}
    p(y)
    =
    \mathbb{E}_{x\sim p_\theta}
    \left[
        p(y\mid x)
    \right]
\end{equation}
be the marginal label distribution over generated samples.
The Inception Score is defined as
\begin{equation}
    \mathrm{IS}
    =
    \exp
    \left(
        \mathbb{E}_{x\sim p_\theta}
        \left[
            \mathrm{KL}
            \bigl(
                p(y\mid x)
                \,\|\,
                p(y)
            \bigr)
        \right]
    \right).
\end{equation}
A higher IS indicates that individual generated images are confidently classified while the overall generated set covers diverse semantic classes.
In our experiments, IS is computed using \(10\) splits, and we report the mean and standard deviation across splits.

%%%%%%%%%%%%%%%%%%%%%%%%%%%%%%%%%%%%%%%%%%%%%%%%%%%%%%%%%%%%
\paragraph{Uncertainty reporting.}
For image-generation metrics, we report KID and Inception Score as mean $\pm$ standard deviation following the evaluation protocol used by the metric implementation, where the standard deviation is computed over the corresponding evaluation splits. FID is reported as a single aggregate estimate. We do not claim statistical significance across independently trained seeds; for single-run qualitative analyses, such as trajectory and field visualizations, the results are used as diagnostic evidence rather than statistical significance claims.

%%%%%%%%%%%%%%%%%%%%%%%%%%%%%%%%%%%%%%%%%%%%%%%%%%%%%%%%%%%%
\subsection{CIFAR-10}
\label{app:cifar10}

\begin{table}[h]
\centering
\caption{Training configuration for CIFAR-10 image experiments.}
\label{tab:app_cifar10_training_config}
\begin{tabular}{ll}
\toprule
Hyperparameter & Value \\
\midrule
Dataset & CIFAR-10 \\
Batch size & \(128\) \\
Epochs & \(300\) \\
Gradient accumulation & \(1\) \\
Learning rate & \(10^{-4}\) \\
Optimizer betas & \((0.9, 0.95)\) \\
Learning-rate decay & Disabled \\
EMA & Enabled \\
Class drop probability & \(1.0\) \\
CFG scale & \(0.0\) \\
Skewed timestep sampling & Enabled \\
EDM schedule & Enabled \\
Sampling solver & Heun2 \\
Number of function evaluations & \(50\) \\
Sampling dtype & float32 \\
FID samples & \(50{,}000\) \\
Evaluation frequency & \(100\) epochs \\
Checkpoint frequency & \(20\) epochs \\
Number of workers & \(10\) \\
Pin memory & Enabled \\
Distributed training & Enabled \\
Number of GPUs & \(2\) \\
Distributed backend & NCCL \\
Seed & \(0\) \\
\bottomrule
\end{tabular}
\end{table}

\begin{table}[h]
\centering
\caption{Method-specific settings for CIFAR-10 CFM-Diffusion and CBM-Diffusion.}
\label{tab:app_cifar10_method_config}
\begin{tabular}{lll}
\toprule
Setting & CFM-Diffusion & CBM-Diffusion \\
\midrule
Target type & \texttt{cfm\_diffusion} & \texttt{cbm\_diffusion} \\
\(\beta_{\mathrm{impl}}\) & \(0.01\) & \(0.01\) \\
\(\sigma_{\min}\) & \(0.05\) & \(0.05\) \\
\(t_{\epsilon}\) & \(10^{-4}\) & \(10^{-4}\) \\
\(\lambda_d\) & \(1.0\) & \(1.0\) \\
Forward alignment weight & \(0.0\) & \(0.0\) \\
Posterior temperature & \(1.0\) & \(1.0\) \\
Posterior top-\(k\) & Disabled & Disabled \\
VP scheduler \(\beta_{\min}\) & \(0.1\) & \(0.1\) \\
VP scheduler \(\beta_{\max}\) & \(20.0\) & \(20.0\) \\
\bottomrule
\end{tabular}
\end{table}

\begin{table}[h]
\centering
\caption{Evaluation configuration for CIFAR-10 image generation.}
\label{tab:app_cifar10_eval_config}
\begin{tabular}{ll}
\toprule
Setting & Value \\
\midrule
Dataset & CIFAR-10 \\
Checkpoint & Epoch \(299\) \\
Evaluation mode & \texttt{eval\_only} \\
Evaluation batch size & \(512\) \\
Sampling solver & Heun2 \\
Number of function evaluations & \(50\) \\
Sampling dtype & float32 \\
Generated samples & \(50{,}000\) \\
FID & Enabled \\
KID & Enabled \\
Inception Score & Enabled \\
Inception splits & \(10\) \\
Classifier-free guidance scale & \(0.0\) \\
Class drop probability & \(1.0\) \\
EMA checkpoint & Used \\
Number of workers & \(10\) \\
Pin memory & Enabled \\
Distributed evaluation & Enabled \\
Number of GPUs & \(2\) \\
Distributed backend & NCCL \\
Seed & \(0\) \\
\bottomrule
\end{tabular}
\end{table}

CIFAR-10 consists of \(60{,}000\) color images with spatial resolution \(32\times32\), divided into \(50{,}000\) training images and \(10{,}000\) test images.
The dataset contains \(10\) object categories, with \(6{,}000\) images per class.
In our experiments, we use CIFAR-10 as a standard low-resolution natural image benchmark.

\paragraph{Training configuration.}
For CIFAR-10 image generation, we compare CFM-Diffusion and CBM-Diffusion under the same experimental protocol.
Table~\ref{tab:app_cifar10_training_config} summarizes the main training arguments that are explicitly set for this experiment, while Table~\ref{tab:app_cifar10_method_config} reports the method-specific target and bridge parameters.
All arguments not shown in these tables use the default values from the training parser, and the complete parsed configuration is saved as \texttt{args.json} in the output directory of each run.

Both methods use the same architecture, optimizer, data pipeline, timestep sampling strategy, EMA setting, and sampling configuration.
The only methodological difference is the target construction:
CFM-Diffusion uses the standard diffusion Flow Matching target, while CBM-Diffusion decomposes the same diffusion-path target into transport and osmotic components.
For CFM-Diffusion, the target type is set to
\[
\texttt{cfm\_diffusion},
\]
where the model learns the full diffusion-path velocity field directly.
For CBM-Diffusion, the target type is set to
\[
\texttt{cbm\_diffusion},
\]
where the target velocity is decomposed into \(u_t^*\) and \(d_t^*\), and the model is trained with the Bridge Matching objective.

\paragraph{Evaluation configuration.}
For CIFAR-10, we evaluate the final EMA checkpoint after training.
Specifically, we use the checkpoint at epoch \(299\) and run evaluation in \texttt{eval\_only} mode.
Table~\ref{tab:app_cifar10_eval_config} summarizes the main evaluation arguments used in the reported experiment.
As with training, unspecified evaluation arguments follow the default values from the evaluation parser, and the full parsed evaluation configuration is stored in the corresponding \texttt{args.json} file.

Generated samples are obtained by solving the learned probability-flow ODE with the Heun2 solver using \(50\) number of function evaluations.
For CBM-Diffusion, the learned fields are recombined during sampling as
\begin{equation}
    v_{\theta,\phi}^{(\lambda_u,\lambda_d)}(x,t)
    =
    \lambda_u u_\theta(x,t)
    +
    \lambda_d d_\phi(x,t),
\end{equation}
where \((\lambda_u,\lambda_d)\) is specified for each evaluation setting.
For the CFM-Diffusion baseline, the same evaluation interface is used with the learned velocity field evaluated directly.
%%%%%%%%%%%%%%%%%%%%%%%%%%%%%%%%%%%%%%%%%%%%%%%%%%%%%%%%%%%%
\subsection{ImageNet-32}
\label{app:imagenet32}
\begin{table}[h]
\centering
\caption{Training configuration for ImageNet-32 experiments.}
\label{tab:app_imagenet32_training_config}
\begin{tabular}{ll}
\toprule
Setting & Value \\
\midrule
Dataset & ImageNet-32 \\
Batch size & \(128\) \\
Epochs & \(1000\) \\
Gradient accumulation & \(2\) \\
Learning rate & \(10^{-5}\) \\
Optimizer betas & \((0.9,0.95)\) \\
Learning-rate decay & Disabled \\
EMA & Enabled \\
Class drop probability & \(0.2\) \\
Skewed timestep sampling & Enabled \\
EDM schedule & Enabled \\
Sampling solver during training eval & Heun2 \\
Training-eval NFE & \(50\) \\
Sampling dtype & float32 \\
CFG scale & \(2.0\) \\
FID samples during training eval & \(10{,}000\) \\
Evaluation frequency & \(200\) epochs \\
Checkpoint frequency & \(50\) epochs \\
Number of workers & \(10\) \\
Pin memory & Enabled \\
Distributed training & Enabled \\
Number of GPUs & \(1\) \\
Distributed backend & NCCL \\
Seed & \(0\) \\
\bottomrule
\end{tabular}
\end{table}
\begin{table}[h]
\centering
\caption{Final evaluation configuration for ImageNet-32 experiments.}
\label{tab:app_imagenet32_eval_config}
\begin{tabular}{ll}
\toprule
Setting & Value \\
\midrule
Dataset & ImageNet-32 \\
Checkpoint & Epoch \(999\) \\
Evaluation mode & \texttt{eval\_only} \\
Evaluation batch size & \(256\) \\
Sampling solver & Heun2 \\
Number of function evaluations & \(100\) \\
Sampling dtype & float32 \\
Generated samples & \(10{,}000\) \\
FID & Enabled \\
KID & Enabled \\
Inception Score & Enabled \\
Inception splits & \(10\) \\
CFG scale & \(2.0\) \\
Class drop probability & \(0.2\) \\
EMA checkpoint & Used \\
Number of workers & \(10\) \\
Pin memory & Enabled \\
Distributed evaluation & Enabled \\
Number of GPUs & \(1\) \\
Distributed backend & NCCL \\
Seed & \(0\) \\
\bottomrule
\end{tabular}
\end{table}

ImageNet-32 is a downsampled version of ImageNet, where each image is resized to \(32\times32\) resolution.
Compared with CIFAR-10, ImageNet-32 contains substantially more semantic diversity and a larger number of object categories, making it a more challenging benchmark at the same spatial resolution.
We use ImageNet-32 to evaluate whether the proposed decomposition remains effective beyond small-scale class-balanced datasets.

\paragraph{Training configuration.}
Both CFM-Diffusion and CBM-Diffusion are trained for \(1000\) epochs with batch size \(128\) and gradient accumulation \(2\).
Table~\ref{tab:app_imagenet32_training_config} summarizes the main explicitly specified training arguments for this experiment.
All unlisted arguments use the default values from the training parser, with the complete parsed configuration saved to \texttt{args.json} in the output directory.

We use AdamW with learning rate \(10^{-5}\) and optimizer betas \((0.9,0.95)\).
Learning-rate decay is disabled, and EMA is enabled during training.
The class drop probability is set to \(0.2\), and classifier-free guidance is used at sampling time with guidance scale \(2.0\).
We use skewed timestep sampling and the EDM schedule.
The same configuration is used for CFM-Diffusion and CBM-Diffusion unless a method-specific setting is explicitly stated.

\paragraph{Evaluation configuration.}
For final evaluation, we use the checkpoint at epoch \(999\) and run evaluation in \texttt{eval\_only} mode.
Table~\ref{tab:app_imagenet32_eval_config} summarizes the main explicitly specified evaluation arguments.
All unspecified evaluation arguments follow the default values from the evaluation parser, and the complete parsed configuration is saved in \texttt{args.json}.

Compared with the training-time evaluation, we increase the number of function evaluations from \(50\) to \(100\).
We generate \(10{,}000\) samples and compute FID, KID, and IS.
KID and IS are computed using \(10\) splits.
The evaluation batch size is \(256\), and sampling is performed in float32 using the Heun2 solver.
%%%%%%%%%%%%%%%%%%%%%%%%%%%%%%%%%%%%%%%%%%%%%%%%%%%%%%%%%%%%
\subsection{ImageNet-64}
\label{app:imagenet64}

\begin{table}[h]
\centering
\caption{Training configuration for ImageNet-64 experiments.}
\label{tab:app_imagenet64_training_config}
\begin{tabular}{ll}
\toprule
Setting & Value \\
\midrule
Dataset & ImageNet-64 \\
Batch size & \(8\) \\
Gradient accumulation & \(8\) \\
Effective batch size & \(64\) \\
Epochs & \(500\) \\
Learning rate & \(10^{-5}\) \\
Optimizer betas & \((0.9,0.95)\) \\
Learning-rate decay & Disabled \\
EMA & Enabled \\
Class drop probability & \(0.2\) \\
Skewed timestep sampling & Enabled \\
EDM schedule & Enabled \\
Sampling solver during training eval & Heun2 \\
Training-eval NFE & \(50\) \\
Sampling dtype & float32 \\
CFG scale & \(2.0\) \\
FID samples during training eval & \(10{,}000\) \\
Evaluation frequency & \(100\) epochs \\
Checkpoint frequency & \(50\) epochs \\
Number of workers & \(10\) \\
Pin memory & Enabled \\
Distributed training & Enabled \\
Number of GPUs & \(1\) \\
Distributed backend & NCCL \\
Seed & \(0\) \\
\bottomrule
\end{tabular}
\end{table}

\begin{table}[h]
\centering
\caption{Evaluation configuration for ImageNet-64 experiments.}
\label{tab:app_imagenet64_eval_config}
\begin{tabular}{ll}
\toprule
Setting & Value \\
\midrule
Dataset & ImageNet-64 \\
Evaluation mode & \texttt{eval\_only} \\
Evaluation batch size & \(8\) \\
Gradient accumulation & \(8\) \\
Sampling solver & Heun2 \\
Number of function evaluations & \(50\) \\
Sampling dtype & float32 \\
Generated samples & \(10{,}000\) \\
FID & Enabled \\
KID & Disabled \\
Inception Score & Disabled \\
CFG scale & \(2.0\) \\
Class drop probability & \(0.2\) \\
EMA checkpoint & Used \\
Number of workers & \(10\) \\
Pin memory & Enabled \\
Device & CUDA \\
Distributed backend & NCCL \\
Seed & \(0\) \\
\(\beta_{\mathrm{impl}}\) & \(0.01\) \\
\(\sigma_{\min}\) & \(0.05\) \\
\(t_\epsilon\) & \(10^{-4}\) \\
VP scheduler \(\beta_{\min}\) & \(0.1\) \\
VP scheduler \(\beta_{\max}\) & \(20.0\) \\
\bottomrule
\end{tabular}
\end{table}

ImageNet-64 is a downsampled ImageNet benchmark with images resized to \(64\times64\) resolution.
It preserves more spatial detail than ImageNet-32 and therefore provides a more challenging setting for generation quality and sampling stability.
We use ImageNet-64 to test the scalability of the proposed method to higher-resolution image generation.

\paragraph{Training configuration.}
For ImageNet-64 experiments, we use the same diffusion-path setup as in the ImageNet-32 experiments, but train at \(64\times64\) resolution.
Table~\ref{tab:app_imagenet64_training_config} summarizes the main explicitly specified training arguments.
All other arguments follow the default values from the training parser, and the complete parsed configuration is saved to \texttt{args.json} in the output directory.

CFM-Diffusion and CBM-Diffusion are trained under the same configuration, with the only difference being the target construction.
CFM-Diffusion learns the standard diffusion Flow Matching target, while CBM-Diffusion decomposes the diffusion-path target into transport and osmotic components.
Due to the higher spatial resolution, we use a smaller per-step batch size and gradient accumulation.
Specifically, we train with batch size \(8\), gradient accumulation \(8\), and effective batch size \(64\).
The model is trained for \(500\) epochs using AdamW with learning rate \(10^{-5}\) and optimizer betas \((0.9,0.95)\).
Learning-rate decay is disabled, and EMA is enabled.

\paragraph{Evaluation configuration.}
For ImageNet-64 evaluation, we use the same evaluation protocol as ImageNet-32 unless otherwise stated.
Evaluation is performed in \texttt{eval\_only} mode by resuming from the trained checkpoint.
Table~\ref{tab:app_imagenet64_eval_config} summarizes the main explicitly specified evaluation arguments.
All unlisted evaluation arguments follow the default values from the evaluation parser, and the complete parsed configuration is saved to \texttt{args.json}.

Generated samples are obtained using the Heun2 ODE solver with \(50\) number of function evaluations.
We generate \(10{,}000\) samples and compute FID.
KID and IS are not enabled in this evaluation configuration unless explicitly stated.

All other unmentioned arguments follow the default evaluation-parser values, except where they are explicitly shared with the ImageNet-32 setting.
In particular, we keep the same timestep schedule, EMA usage, classifier-free guidance setting, data-loading options, and diffusion-path scheduler parameters unless otherwise specified.
%%%%%%%%%%%%%%%%%%%%%%%%%%%%%%%%%%%%%%%%%%%%%%%%%%%%%%%%%%%%
%%%%%%%%%%%%%%%%%%%%%%%%%%%%%%%%%%%%%%%%%%%%%%%%%%%%%%%%%%%%
\section{Additional experimental results}
\label{app:additional-results}

%%%%%%%%%%%%%%%%%%%%%%%%%%%%%%%%%%%%%%%%%%%%%%%%%%%%%%%%%%%%%%%%%%
\subsection{Additional results on 2D transport tasks}

\label{app:additional-2d-results}

\begin{figure}[p]
    \centering
    \includegraphics[width=\textwidth]{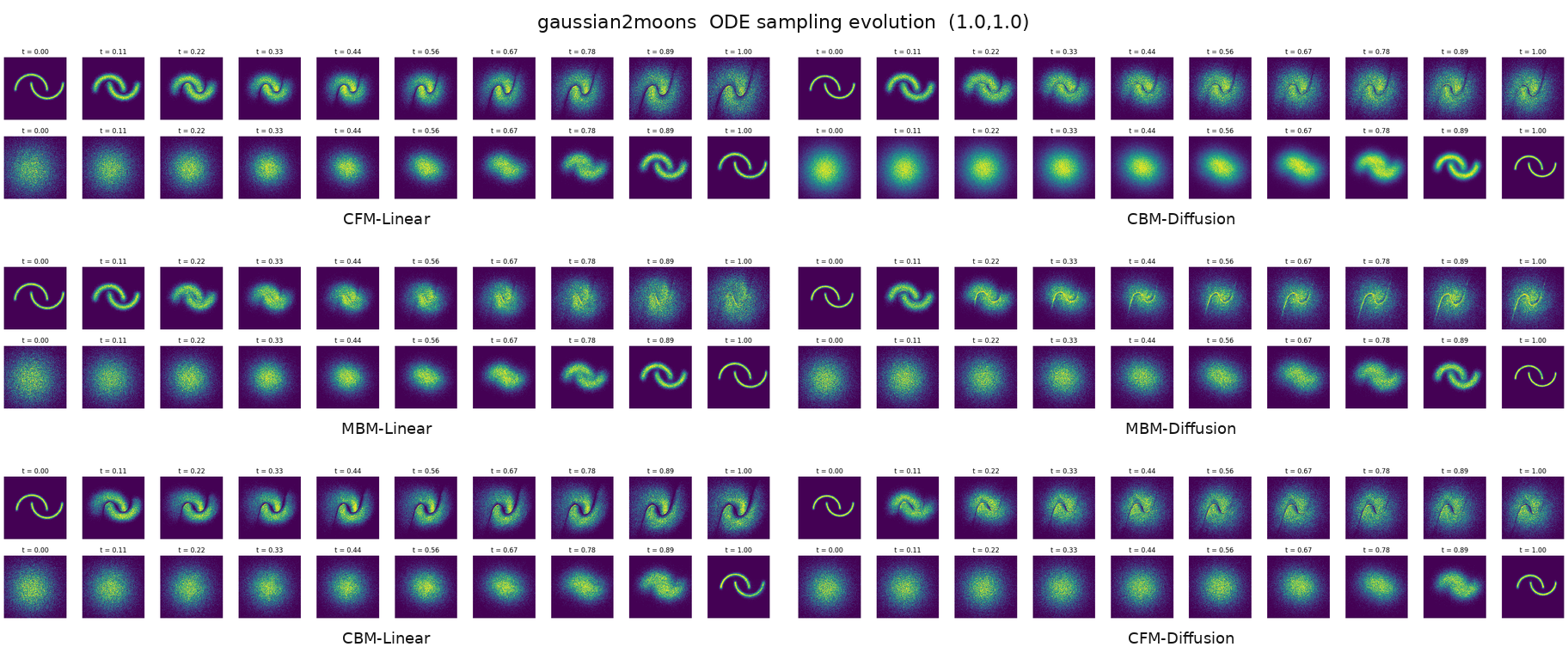}
    \caption{
    ODE sampling evolution on the Gaussian-to-moons transport task.
    We compare CFM-Linear, CFM-Diffusion, CBM-Linear, CBM-Diffusion, MBM-Linear, and MBM-Diffusion under \((\lambda_u,\lambda_d)=(1.0,1.0)\).
    }
    \label{fig:app_gaussian2moons_transport}
\end{figure}

\begin{figure}[p]
    \centering
    \includegraphics[width=\textwidth]{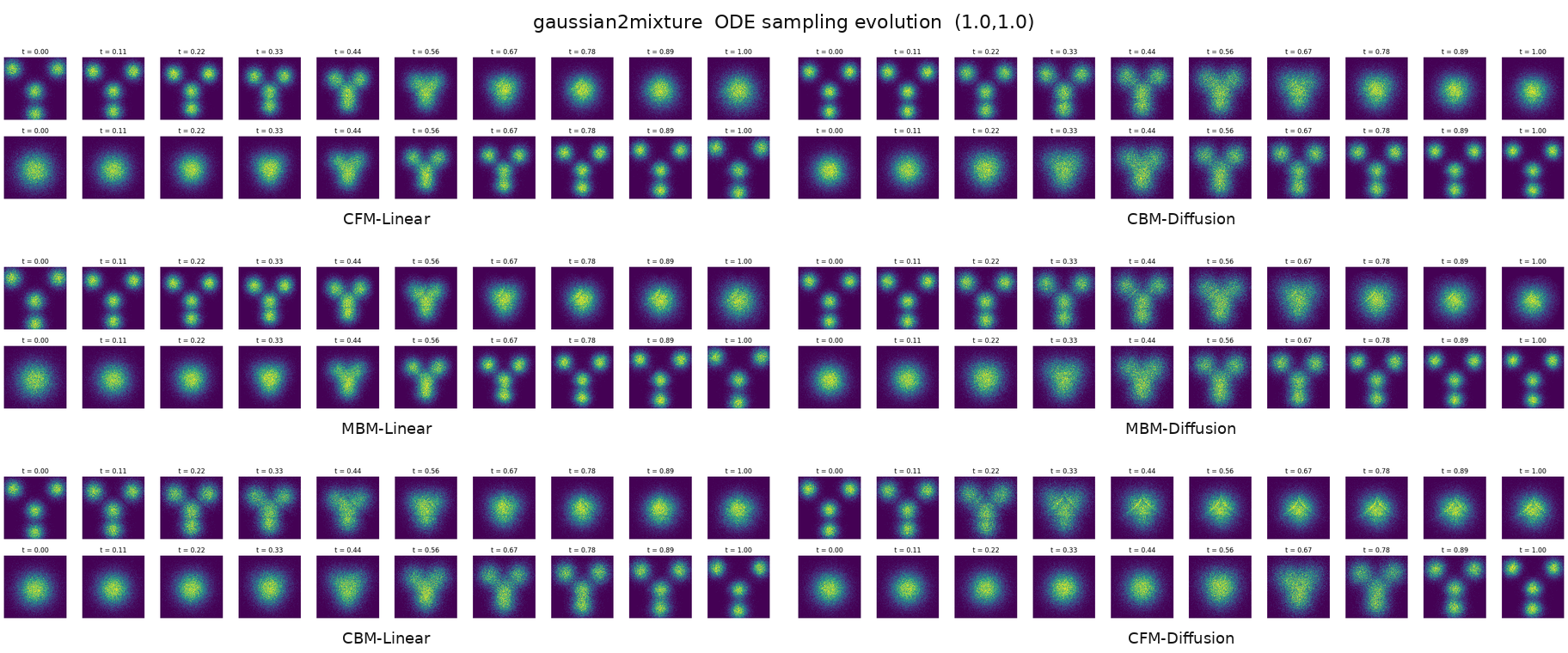}
    \caption{
    ODE sampling evolution on the Gaussian-to-mixture transport task.
    We compare CFM-Linear, CFM-Diffusion, CBM-Linear, CBM-Diffusion, MBM-Linear, and MBM-Diffusion under \((\lambda_u,\lambda_d)=(1.0,1.0)\).
    }
    \label{fig:app_gaussian2mixture_transport}
\end{figure}

\begin{figure}[p]
    \centering
    \includegraphics[width=\textwidth]{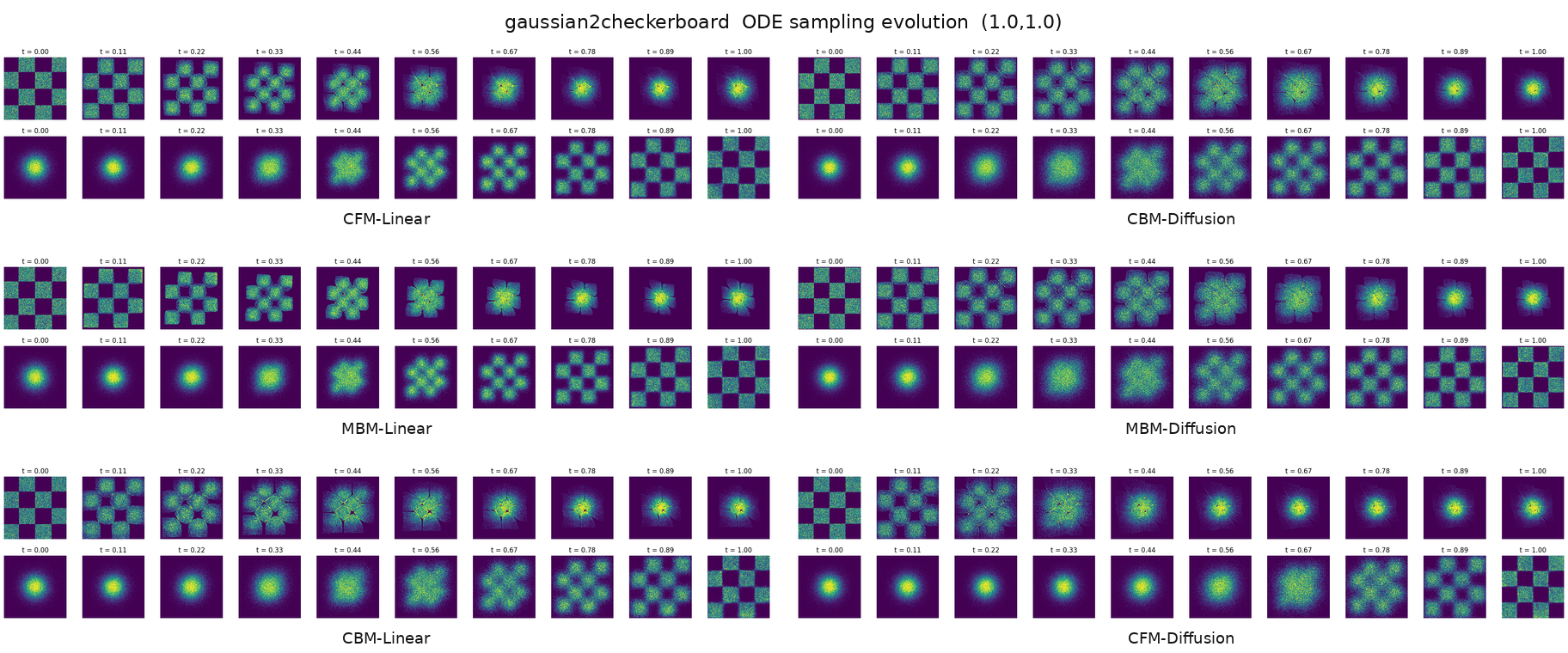}
    \caption{
    ODE sampling evolution on the Gaussian-to-checkerboard transport task.
    We compare CFM-Linear, CFM-Diffusion, CBM-Linear, CBM-Diffusion, MBM-Linear, and MBM-Diffusion under \((\lambda_u,\lambda_d)=(1.0,1.0)\).
    }
    \label{fig:app_gaussian2checkerboard_transport}
\end{figure}

\begin{figure}[p]
    \centering
    \includegraphics[width=\textwidth]{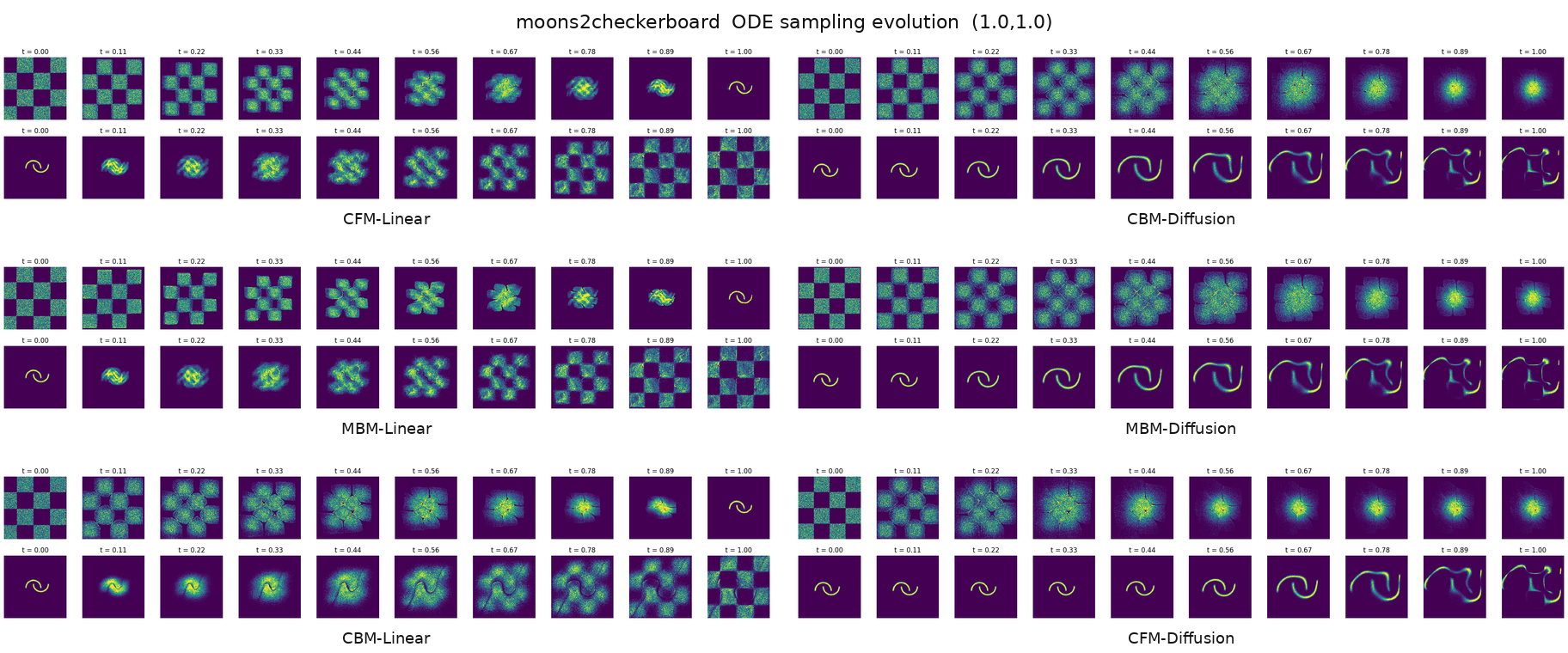}
    \caption{
    ODE sampling evolution on the moons-to-checkerboard transport task.
    We compare CFM-Linear, CFM-Diffusion, CBM-Linear, CBM-Diffusion, MBM-Linear, and MBM-Diffusion under \((\lambda_u,\lambda_d)=(1.0,1.0)\).
    }
    \label{fig:app_moons2checkerboard_transport}
\end{figure}

\begin{figure}[p]
    \centering
    \includegraphics[width=\textwidth]{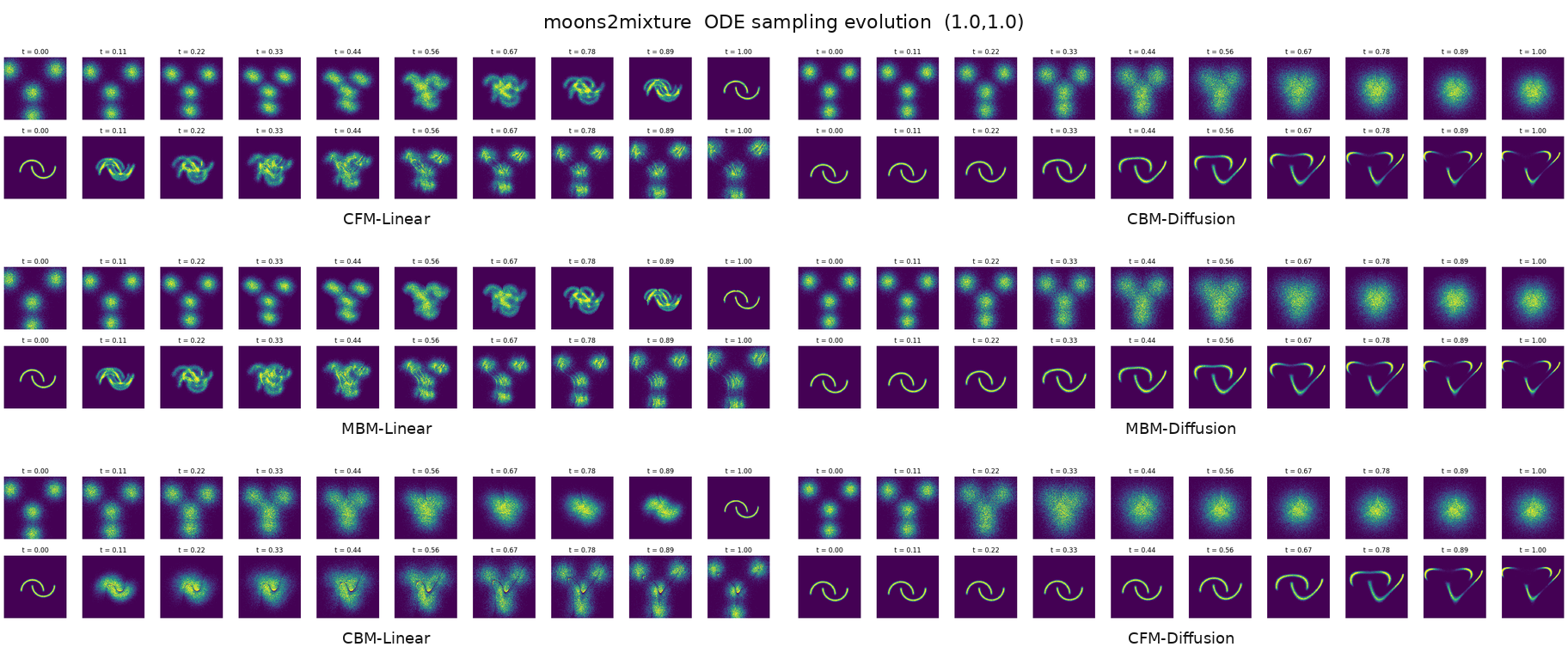}
    \caption{
    ODE sampling evolution on the moons-to-mixture transport task.
    We compare CFM-Linear, CFM-Diffusion, CBM-Linear, CBM-Diffusion, MBM-Linear, and MBM-Diffusion under \((\lambda_u,\lambda_d)=(1.0,1.0)\).
    }
    \label{fig:app_moons2mixture_transport}
\end{figure}

\begin{figure}[p]
    \centering
    \includegraphics[width=\textwidth]{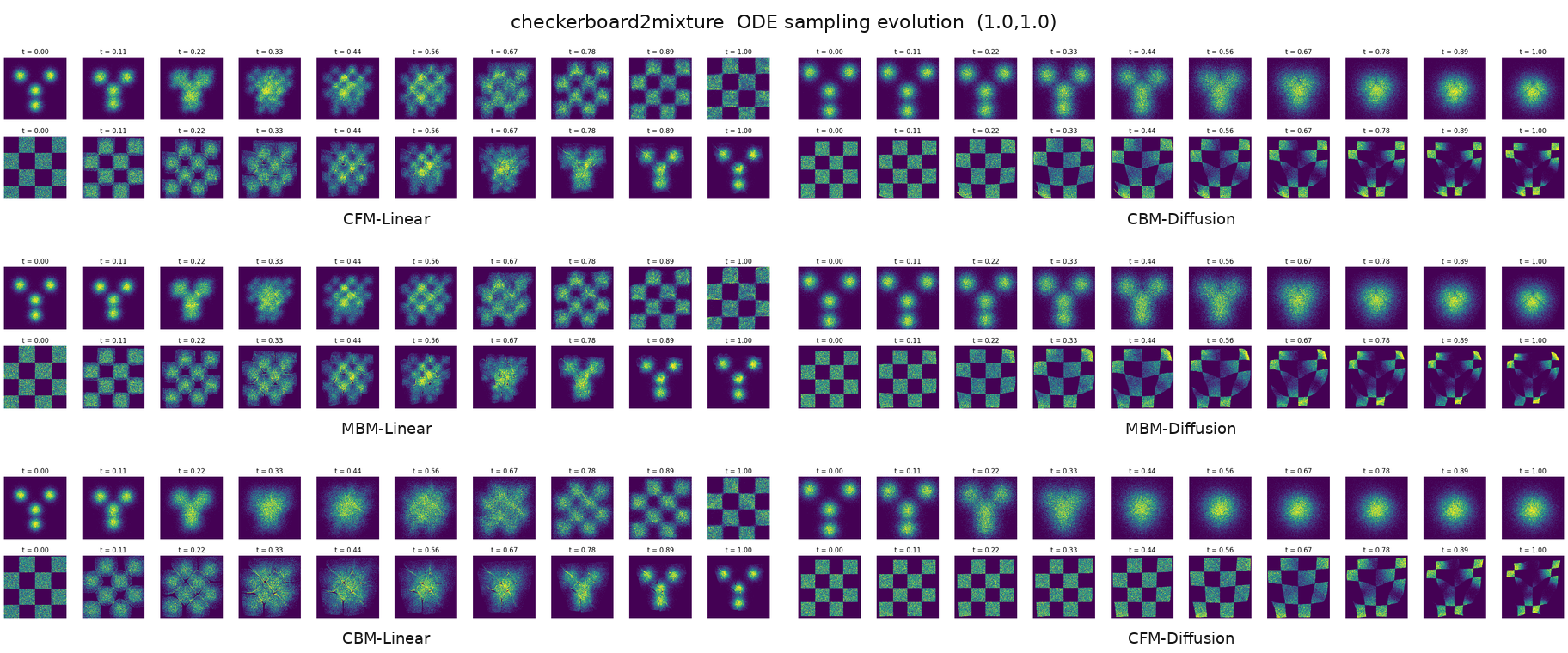}
    \caption{
    ODE sampling evolution on the checkerboard-to-mixture transport task.
    We compare CFM-Linear, CFM-Diffusion, CBM-Linear, CBM-Diffusion, MBM-Linear, and MBM-Diffusion under \((\lambda_u,\lambda_d)=(1.0,1.0)\).
    }
    \label{fig:app_checkerboard2mixture_transport}
\end{figure}

We provide additional qualitative and quantitative results on 2D transport tasks. shown in Fig. \ref{fig:app_gaussian2moons_transport} to \ref{fig:app_checkerboard2mixture_transport}.
For each source--target pair, we visualize the ODE sampling evolution of six methods:
CFM-Linear, CFM-Diffusion, CBM-Linear, CBM-Diffusion, MBM-Linear, and MBM-Diffusion.
All visualizations are generated using the recombined dynamics with \((\lambda_u,\lambda_d)=(1.0,1.0)\).
Each panel shows the evolution of generated samples along the learned ODE trajectory from \(t=0\) to \(t=1\).

%%%%%%%%%%%%%%%%%%%%%%%%%%%%%%%%%%%%%%%%%%%%%%%%%%%%%%%%%%%%%%%%%%
\subsection{Additional 2D ablation on the osmotic weight}

\label{app:additional-2d-lambda-ablation}

\begin{figure}[h]
    \centering
    \includegraphics[width=\textwidth]{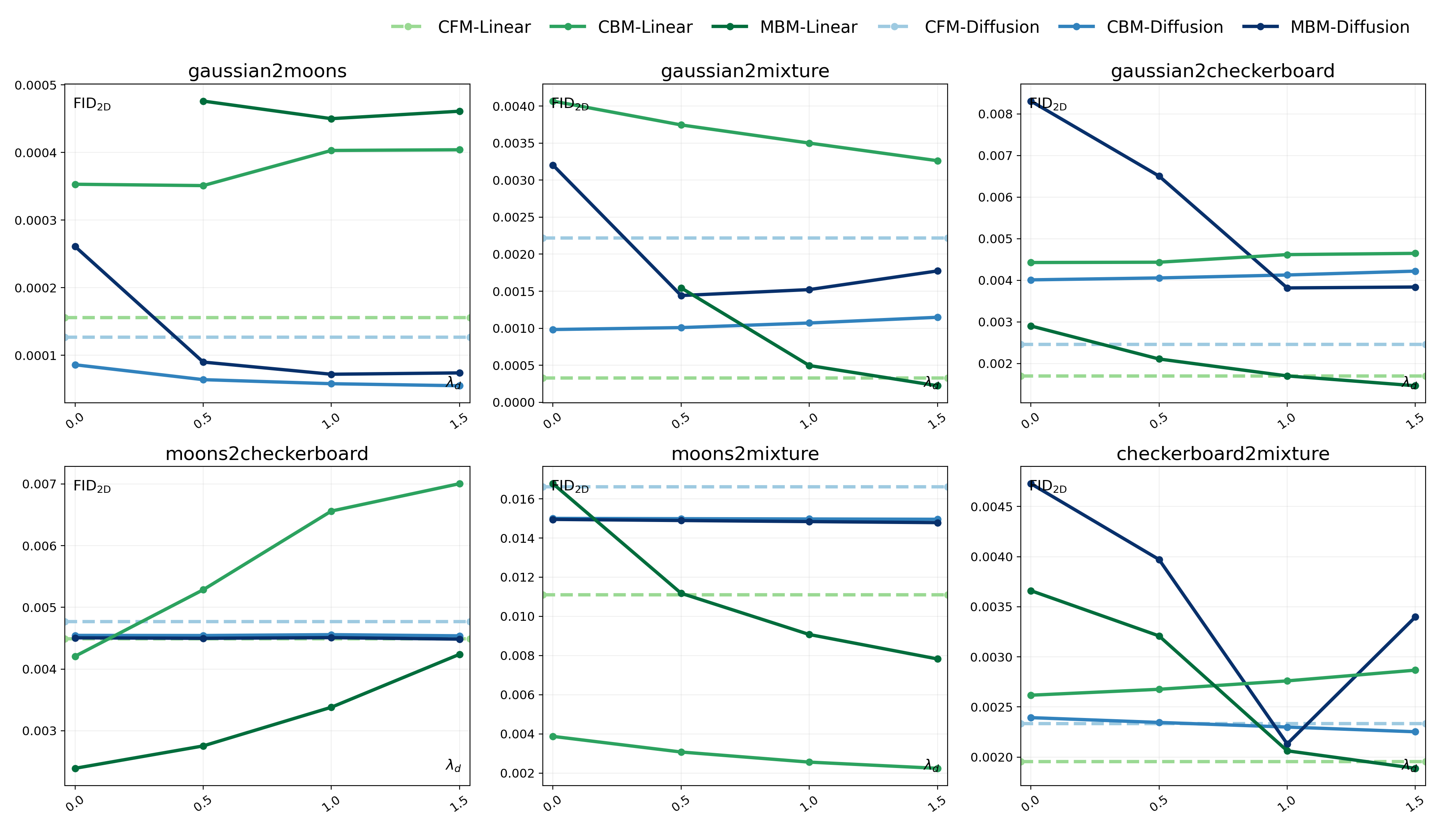}
    \caption{
    Additional 2D ablation on the osmotic recombination weight \(\lambda_d\).
    We report \(\mathrm{FID}_{2D}\) across multiple source--target pairs while fixing \(\lambda_u=1.0\).
    Dashed lines correspond to CFM baselines, and solid lines correspond to BM variants evaluated with different \(\lambda_d\) values.
    }
    \label{fig:app_all_pairs_fid_lambda_ablation}
\end{figure}

We further evaluate the effect of the osmotic recombination weight \(\lambda_d\) on 2D generation performance, shown in Fig. \ref{fig:app_all_pairs_fid_lambda_ablation}.
For all BM variants, we fix \(\lambda_u=1.0\) and vary \(\lambda_d\in\{0.0,0.5,1.0,1.5\}\).
The dashed horizontal lines indicate the corresponding CFM baselines, while solid curves show the BM variants.

%%%%%%%%%%%%%%%%%%%%%%%%%%%%%%%%%%%%%%%%%%%%%%%%%%%%%%%%%%%%%%%%%%
\subsection{Additional image generation samples}

\label{app:additional-image-samples}

\begin{figure}[h]
    \centering
    \includegraphics[width=0.6\textwidth]{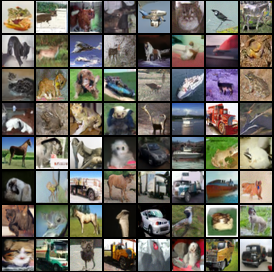}
    \caption{
    Additional generated samples on CIFAR-10 from the checkpoint at epoch \(299\).
    }
    \label{fig:app_cifar10_samples}
\end{figure}

\begin{figure}[h]
    \centering
    \includegraphics[width=0.6\textwidth]{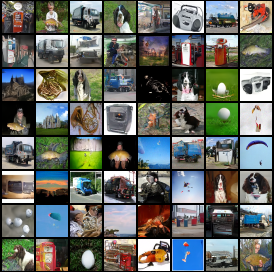}
    \caption{
    Additional generated samples on ImageNet-32 from the checkpoint at epoch \(999\).
    }
    \label{fig:app_imagenet32_samples}
\end{figure}

\begin{figure}[p]
    \centering
    \includegraphics[width=\textwidth]{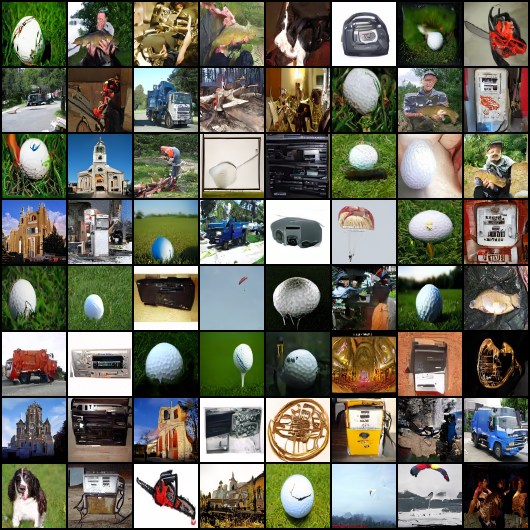}
    \caption{
    Additional generated samples on ImageNet-64 from the checkpoint at epoch \(499\).
    }
    \label{fig:app_imagenet64_samples}
\end{figure}

We provide additional generated samples from the image-generation experiments.
The samples are generated from the final evaluated checkpoints for CIFAR-10 (Fig.\ref{fig:app_cifar10_samples}), ImageNet-32 (Fig. \ref{fig:app_imagenet32_samples}), and ImageNet-64 (Fig. \ref{fig:app_imagenet64_samples}).

%%%%%%%%%%%%%%%%%%%%%%%%%%%%%%%%%%%%%%%%%%%%%%%%%%%%%%%%%%%%%%%%%%

\subsection{Additional sampling-time controllability results}

\label{app:additional-sampling-controllability}

\begin{figure}[t]
    \centering
    \includegraphics[width=\linewidth]{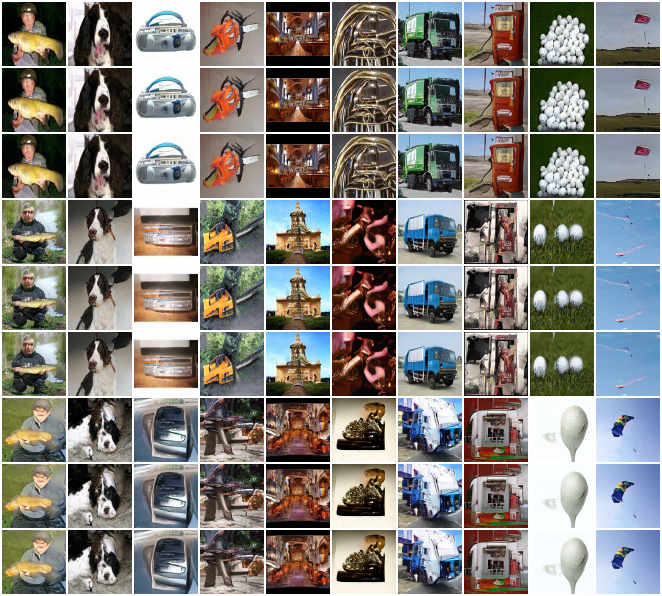}
    \caption{
    Additional sampling-time controllability results obtained by varying the osmotic recombination weight \(\lambda_d\).
    Samples are grouped by class, where each group illustrates how adjusting the strength of the osmotic component changes fine-grained visual appearance while largely preserving semantic content.
    }
    \label{fig:additional_lambda_d_triplets}
\end{figure}

Figure~\ref{fig:additional_lambda_d_triplets} provides additional qualitative results for sampling-time controllability under different osmotic recombination strengths.
For each class, we visualize generated samples obtained by varying the contribution of the learned osmotic field \(d_t\) while keeping the transport component \(u_t\) fixed.
The results further show that changing \(\lambda_d\) affects visual attributes such as texture, shape sharpness, and local density-dependent details, while preserving the overall semantic structure of the generated samples.
This supports the main observation that the transport--osmotic decomposition provides an interpretable sampling-time control mechanism beyond the 2D toy setting.

%%%%%%%%%%%%%%%%%%%%%%%%%%%%%%%%%%%%%%%%%%%%%%%%%%%%%%%%%%%%%%%%%%
%%%%%%%%%%%%%%%%%%%%%%%%%%%%%%%%%%%%%%%%%%%%%%%%%%%%%%%%%%%%%%%%%%
\section{Code availability and implementation resources}

\label{app:code_availability}

To support reproducibility, we include the implementation of our method as part of the supplementary material submitted with this paper.
The released code contains the training and evaluation scripts used for the 2D experiments and image-generation experiments, together with configuration files and instructions for reproducing the reported results.
Due to the supplementary material size limit, we include source code and configuration files but do not include trained model checkpoints or generated sample archives.

Our implementation follows the general experimental structure of Flow Matching-based generative modeling.
In particular, we refer to the official Flow Matching codebase released with the Flow Matching guide~\citep{lipman2024flowmatchingguide,facebookflowmatching} and the PyTorch Flow Matching implementation of~\citet{shihara2023flowmatchinggithub} as implementation references.
These resources provide standard utilities for Flow Matching training, sampling, and evaluation, while our supplementary code adds the proposed Bridge Matching objectives, the transport--osmotic decomposition, and the sampling-time recombination of learned fields.

%%%%%%%%%%%%%%%%%%%%%%%%%%%%%%%%%%%%%%%%%%%%%%%%%%%%%%%%%%%%%%%%%%
%%%%%%%%%%%%%%%%%%%%%%%%%%%%%%%%%%%%%%%%%%%%%%%%%%%%%%%%%%%%%%%%%%
\section{Broader impacts}

\label{app:broader-impacts}

This work is primarily foundational and aims to improve the interpretability and controllability of stochastic generative modeling.
By decomposing generative dynamics into transport and osmotic components, the proposed framework may help researchers better understand how diffusion-induced effects influence probability transport, intermediate trajectories, and final sample quality.
This could have positive impacts on the design of more transparent and controllable generative models, especially in settings where understanding or adjusting sampling behavior is important.

At the same time, improvements in generative modeling may also contribute to potential negative societal impacts.
More controllable or higher-quality generative models could be misused to synthesize misleading images, support disinformation, or generate content that violates privacy or consent.
Although this paper does not release a large-scale pretrained image generator or introduce a new scraped dataset, the methodological ideas may eventually be incorporated into more capable generative systems.
Responsible deployment of such systems should therefore consider safeguards such as dataset governance, watermarking, provenance tracking, content moderation, and restrictions on malicious use.